\PassOptionsToPackage{section}{placeins} 
\documentclass[]{ailab}

\usepackage[T1]{fontenc}

\usepackage{amsmath,amssymb}
\usepackage{amsthm}
\usepackage{mathtools}
\usepackage{tabularx}
\usepackage{url}
\usepackage{adjustbox}
\usepackage{multirow}
\usepackage{pifont}
\usepackage{colortbl}
\usepackage{booktabs}
\usepackage{natbib}
\setcitestyle{square,comma,numbers,sort&compress}
\usepackage{graphicx}
\usepackage{subcaption}
\usepackage{wrapfig}
\usepackage{tcolorbox}
\RequirePackage{xspace}
\usepackage{nicefrac}
\usepackage{enumitem}
\usepackage{float}
\usepackage[section]{placeins}
\usepackage{algorithm}
\usepackage{algpseudocode}
\usepackage{listings}
\usepackage{tikz}
\usetikzlibrary{positioning,arrows.meta,fit,calc}

\definecolor{darkblue}{rgb}{0, 0, 0.5}
\hypersetup{colorlinks=true, citecolor=darkblue, linkcolor=darkblue, urlcolor=darkblue}

\theoremstyle{plain}
\newtheorem{theorem}{Theorem}[section]

\theoremstyle{definition}
\newtheorem{proposition}[theorem]{Proposition}

\newtcolorbox{takeaway}[1][]{colback=metabg!6,colframe=metabg!70,arc=2.5pt,
  boxrule=0.6pt,left=2mm,right=2mm,top=1.2mm,bottom=1.2mm,#1}

\definecolor{T1Navy}{HTML}{003366}
\definecolor{T1Light}{HTML}{4A7EBB}
\newcommand{\Tonebody}{\color{T1Light}}
\newtcolorbox{keyinsight}[1][]{enhanced,breakable,
  colframe=T1Navy,colbacktitle=T1Navy,coltitle=white,
  colback=T1Navy!5!white,boxrule=0.9pt,arc=2mm,
  left=2.5mm,right=2.5mm,top=1.6mm,bottom=1.6mm,
  fonttitle=\bfseries,#1}

\newtcolorbox[auto counter]{algoboxinner}[2]{enhanced,unbreakable,
  colframe=T1Navy,colbacktitle=T1Navy,coltitle=white,
  colback=T1Navy!5!white,boxrule=0.9pt,arc=2mm,
  left=2.5mm,right=2.5mm,top=1.6mm,bottom=1.6mm,
  fonttitle=\small,
  title={\textbf{Algorithm~\thetcbcounter.} #2},
  label={#1}}
\newenvironment{algobox}[2]
  {\par\addvspace{\medskipamount}%
   \begin{algoboxinner}{#2}{#1}\begin{algorithmic}[1]}
  {\end{algorithmic}\end{algoboxinner}%
   \par\addvspace{\medskipamount}}

\definecolor{T1Gray}{HTML}{4D4D4D}
\newtcolorbox[auto counter]{caseboxinner}[2]{enhanced,unbreakable,
  colframe=T1Gray,colbacktitle=T1Gray,coltitle=white,
  colback=T1Gray!4!white,boxrule=0.7pt,arc=2mm,
  left=2.5mm,right=2.5mm,top=1.4mm,bottom=1.4mm,
  fonttitle=\small,
  title={\textbf{} #2},
  label={#1}}
\newenvironment{casebox}[2]
  {\par\addvspace{\medskipamount}\begin{caseboxinner}{#2}{#1}}
  {\end{caseboxinner}\par\addvspace{\medskipamount}}

\tcbset{breakable}

\makeatletter
\DeclareRobustCommand\onedot{\futurelet\@let@token\@onedot}
\def\@onedot{\ifx\@let@token.\else.\null\fi\xspace}

\def\ie{\emph{i.e}\onedot}

\makeatother

\renewcommand{\paragraph}[1]{\vspace{1.25mm}\noindent\textbf{#1}}
\setlist{itemsep=1pt,topsep=3pt,parsep=0pt}
\definecolor{codegray}{gray}{0.45}
\definecolor{codebg}{gray}{0.96}
\usepackage{amsmath,amsfonts,bm}

\newcommand{\figleft}{{\em (Left)}}

\newcommand{\figright}{{\em (Right)}}
\newcommand{\figtop}{{\em (Top)}}
\newcommand{\figbottom}{{\em (Bottom)}}

\def\eqref#1{equation~\ref{#1}}

\def\1{\bm{1}}

\def\va{{\bm{a}}}

\def\vh{{\bm{h}}}

\def\vm{{\bm{m}}}

\def\vp{{\bm{p}}}
\def\vq{{\bm{q}}}

\def\vs{{\bm{s}}}

\def\vu{{\bm{u}}}
\def\vv{{\bm{v}}}

\def\vx{{\bm{x}}}
\def\vy{{\bm{y}}}

\def\mT{{\bm{T}}}

\DeclareMathAlphabet{\mathsfit}{\encodingdefault}{\sfdefault}{m}{sl}
\SetMathAlphabet{\mathsfit}{bold}{\encodingdefault}{\sfdefault}{bx}{n}
\newcommand{\tens}[1]{\bm{\mathsfit{#1}}}

\def\tR{{\tens{R}}}

\def\gA{{\mathcal{A}}}
\def\gB{{\mathcal{B}}}

\def\gD{{\mathcal{D}}}

\def\gL{{\mathcal{L}}}
\def\gM{{\mathcal{M}}}

\def\gP{{\mathcal{P}}}

\def\gS{{\mathcal{S}}}

\def\gV{{\mathcal{V}}}

\def\sI{{\mathbb{I}}}

\newcommand{\R}{\mathbb{R}}

\newcommand{\KL}{D_{\mathrm{KL}}}
\newcommand{\Var}{\mathrm{Var}}

\usepackage{amsmath,amsfonts,bm}

\def\eqref#1{equation~\ref{#1}}

\def\1{\bm{1}}

\def\va{{\bm{a}}}

\def\vh{{\bm{h}}}

\def\vm{{\bm{m}}}

\def\vp{{\bm{p}}}
\def\vq{{\bm{q}}}

\def\vs{{\bm{s}}}

\def\vu{{\bm{u}}}
\def\vv{{\bm{v}}}

\def\vx{{\bm{x}}}
\def\vy{{\bm{y}}}

\def\mT{{\bm{T}}}

\DeclareMathAlphabet{\mathsfit}{\encodingdefault}{\sfdefault}{m}{sl}
\SetMathAlphabet{\mathsfit}{bold}{\encodingdefault}{\sfdefault}{bx}{n}

\def\tR{{\tens{R}}}

\def\gA{{\mathcal{A}}}
\def\gB{{\mathcal{B}}}

\def\gD{{\mathcal{D}}}

\def\gL{{\mathcal{L}}}
\def\gM{{\mathcal{M}}}

\def\gP{{\mathcal{P}}}

\def\gS{{\mathcal{S}}}

\def\gV{{\mathcal{V}}}

\def\sI{{\mathbb{I}}}

\usepackage{multirow}
\usepackage{diagbox}
\usepackage{makecell}
\usepackage{tabularx}
\usepackage{graphicx}

\usepackage{array}
\usepackage{rotating}

\definecolor{aliceblue}{rgb}{0.94, 0.97, 1.0}
\definecolor{citecolor}{HTML}{0071BC}
\definecolor{linkcolor}{HTML}{ED1C24}
\definecolor{darkgreen}{HTML}{539165}

\makeatletter
\newcommand{\thickhline}{%
 \noalign {\ifnum 0=`}\fi \hrule height 1pt
 \futurelet \reserved@a \@xhline
}
\makeatother

\newcommand{\tablesize}{
  \fontsize{8.6pt}{10pt}\selectfont
}

\newcommand{\Tone}{{\color{metabg}\sffamily\bfseries\slshape T1}\xspace}
\newcommand{\TITO}{TITO\xspace}
\newcommand{\Rthree}{R\textsuperscript{3}\xspace}
\newcommand{\ourmodel}{Qwen3.5-122B-A10B\xspace}

\newcommand{\harbor}{Harbor\xspace}
\newcommand{\terminus}{Terminus-2\xspace}
\newcommand{\sglang}{SGLang\xspace}
\newcommand{\megatron}{Megatron\xspace}
\newcommand{\daytona}{Daytona\xspace}

\newcommand{\tbtwoone}{Terminal-Bench~2.1\xspace}
\newcommand{\tbtwozero}{Terminal-Bench~2.0\xspace}
\newcommand{\lhtb}{Long-Horizon Terminal Bench\xspace}

\newcommand{\tmax}{TMax-15k\xspace}
\newcommand{\rstpool}{RST-38k\xspace}
\newcommand{\strategypool}{T1-15k\xspace}

\newcommand{\code}[1]{\texttt{\small #1}}

\title{\centering \textbf{\Tone}: Terminal Agent Reinforcement Learning for Long-Horizon Tasks}

\author[1,2,*,\dag]{Junyao Yang}
\author[1,*]{Yucheng Shi}
\author[1,3]{Zhongzhi Li}
\author[1,4]{Ruhan Wang}
\author[1,5]{Zongxia Li}
\author[1]{Haitao Mi}
\author[1]{Leowei Liang}

\affiliation[1]{Tencent Hy Foundation Model Frontier}
\affiliation[2]{National University of Singapore}
\affiliation[3]{University of Georgia}
\affiliation[4]{Indiana University}
\affiliation[5]{University of Maryland, College Park}

\contribution[*]{Equal Contribution}
\contribution[\dag]{Corresponding Author}

\email{junyaoyang@u.nus.edu}
\email{tberiusyang@global.tencent.com}
\headercontent{{ \raisebox{-0.15ex}{\includegraphics[height=1.2em]{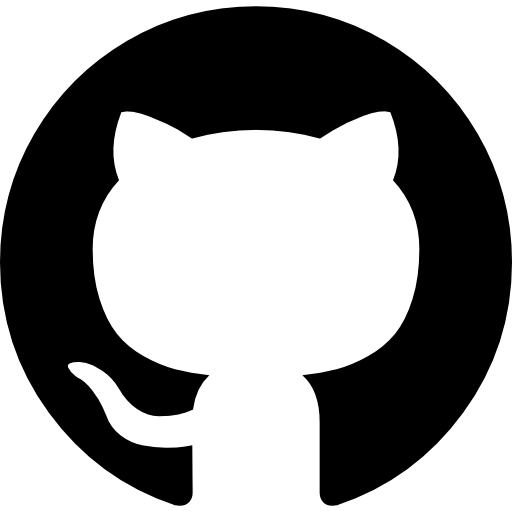}}\hspace{0.5em}\href{https://jyyang26.github.io/t1}{Project Page}}
\qquad
{ {\includegraphics[height=1.2em]{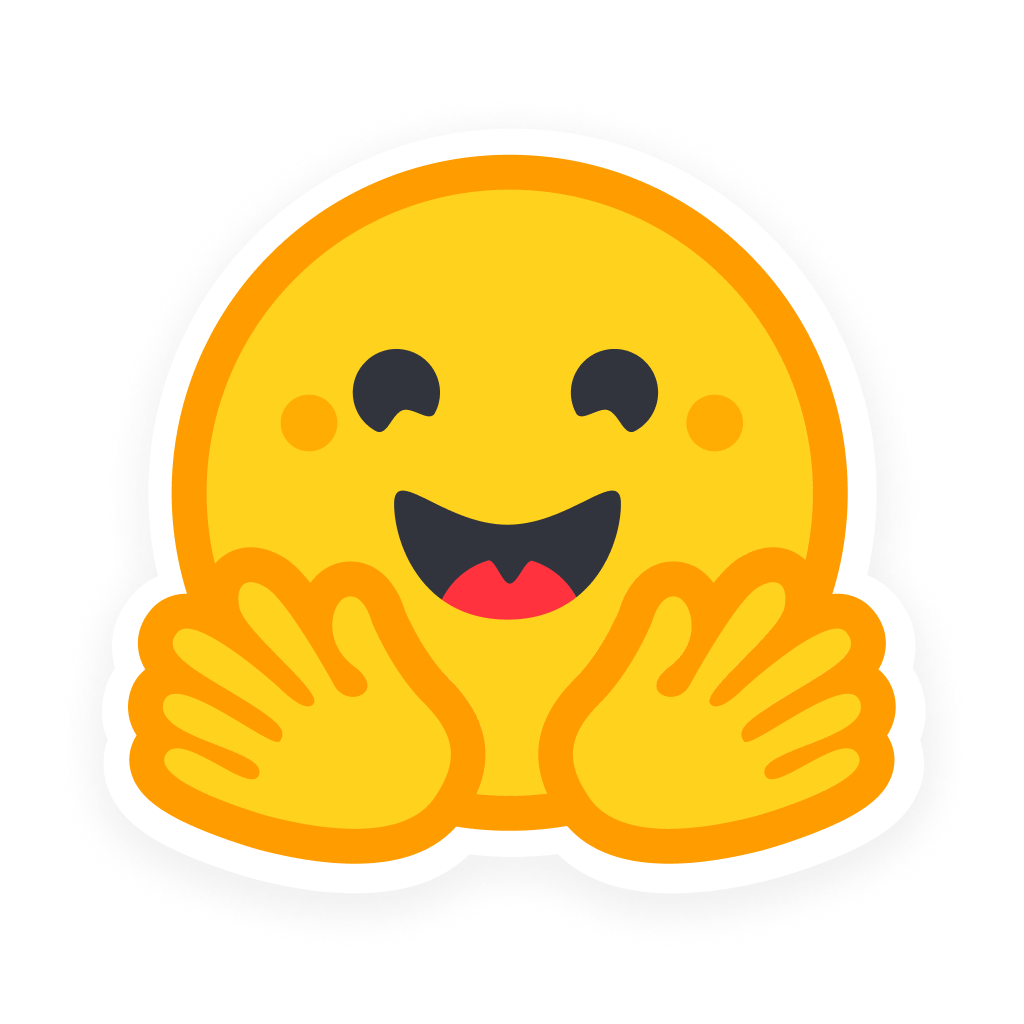}}\hspace{0.5em} \href{https://huggingface.co/collections/TberiusJunyao/t1}{HuggingFace}}}

\abstract{
Agent usage is shifting toward long-horizon tasks such as coding and scientific
discovery, among which terminal tasks are especially important. We introduce \Tone, a
Mixture-of-Experts model of 122B total trained with
reinforcement learning, operating a real shell in a cloud sandbox for up to
\textbf{300+ tool-call turns per task}, rewarded by executing each task's own verifier. 
We provide a comprehensive recipe: First, an aggressively warm-started to stabilize actor-critic training, with a \textbf{dense process reward} scoring
trajectories by the absolute number of passing verifiers. Second, stable optimization
through \textbf{TITO} construction,  training on the
exact sampled token identifiers with drift repair at turn boundaries, and
\textbf{rollout routing replay}, recording the sampler's
per-token expert choices at every MoE layer and replaying them during training. 
Third, \textbf{fully out-of-distribution training corpus}: isolated seeds and synthesized tasks disjoint from \tbtwoone ensures gains reflect genuine capability transfer over benchmark overfitting.
Together, \TITO
and \Rthree cut the training-to-inference log-probability difference from 0.021 to 0.013, with exactly aligned
\textbf{zero token drift} in the loss region. 
On \tbtwoone, our post-train pipeline raises initial base model from 43.8\% to \Tone with \textbf{64.0\% resolved}. On \lhtb, \Tone reaches \textbf{27.9\%} and surpasses GPT-5.4 and GLM-5.1.
\par

{\setlength{\parskip}{0pt}\setlength{\parindent}{0pt}%
 \vspace{2mm}
 \includegraphics[width=\linewidth]{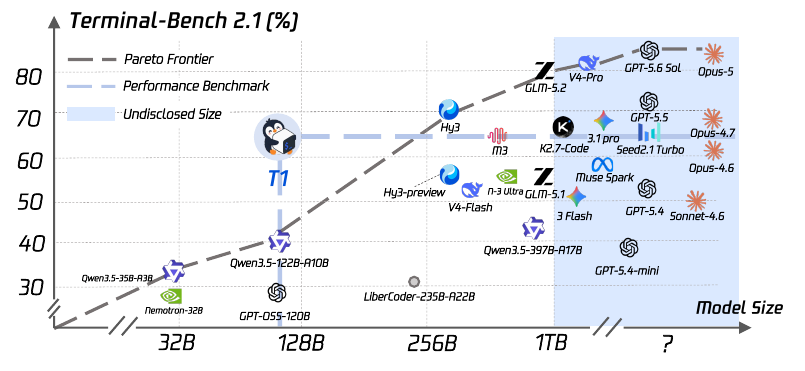}\par
\vspace{1mm}
\captionof{figure}{Performance vs. model size under the same agent harness (\tbtwoone resolved rate). At 122B, \Tone achieves 64.0\%, outperforming GPT-5.4 (54.8\%), DeepSeek-V4-Flash (56.9\%), and Claude Opus 4.6 (63.8\%). It closely approaches Claude Opus 4.7 (66.1\%), and is the best model in its size band.}
 }
}

\date{September 10, 2026}

\begin{document}
\thispagestyle{firstheader}
\maketitle

\tableofcontents
\clearpage

\section{Introduction}
\label{sec:intro}

Large language models have pushed agentic AI from single-turn code completion and
conversational assistance toward the far more demanding domain of autonomous long-horizon
execution \citep{terminalbench}. This marks a critical frontier: a model must no longer
merely produce text satisfying a rubric, but issue actions whose consequences persist in a
stateful environment and withstand verification by execution rather than preference. Among
such environments the \textbf{Linux terminal} is the sharpest and most unforgiving test,
since it binds abstract planning to irreversible side effects and forms the substrate of
nearly all modern software engineering.

Mastering the terminal demands more than command syntax: environment comprehension, task
decomposition, and precise recovery from partial failure, sustained over horizons far
exceeding ordinary reasoning benchmarks. These skills are most rigorously tested by
\textbf{long-horizon terminal benchmarks} such as Long-Horizon Terminal-Bench
\citep{li2026longhorizonterminalbenchtestinglimitsagents} , Terminal-Bench Hard \citep{rst} and Terminal-Bench
\citep{terminalbench}, where one task may require bisecting hundreds of commits, repairing
a defect, rebuilding to a named target and proving the repair, adjudicated by the task's
own held-out verifier. We view reinforcement learning on executed outcomes as the critical
path toward \emph{autonomous software agency}: before models can operate production
systems, reward derived from real execution rather than a learned preference model must be
shown to optimize stably at frontier scale. Terminal performance is thus not the end goal,
but a step toward agents acting on consequential infrastructure.

In this work we introduce \Tone, obtained by post-training \ourmodel \citep{yang2025qwen3}, through
reinforcement learning \citep{ppo,slime} on terminal tasks. Our design confronts
the two difficulties that dominate this regime:

\begin{itemize}[left=0pt]
    \item \textbf{Training-inference consistency for sparse models.}
    Expert weights account for 116.0B of 121.4B parameters, and each token engages 8 of
    256 experts per layer through a discrete router. Minor numeric differences between
    inference and training flip these selections, so gradient may reach different
    parameters from those that generated the behaviour, while multi-turn harnesses perturb
    the token sequence at every turn boundary. We treat these as orthogonal axes, resolved
    separately by \textbf{\TITO} for tokens and \textbf{\Rthree} for experts
    (Sections~\ref{sec:tito} and~\ref{sec:r3}), cutting the measured log-probability gap
    from 0.021 to 0.013 with \textbf{zero drift}
    for training.
    \item \textbf{Dense reward from execution.}
    A rollout batch costs hundreds of sandbox-hours, yet a binary outcome yields one bit
    per trajectory; our first binary-reward campaign never exceeded its supervised
    baseline. We instead score by the \textbf{absolute number of passing assertions} on a
    fixed global scale, feeding a warm-started critic trained at $30\times$ the actor
    learning rate (Sections~\ref{sec:reward} and~\ref{sec:stability}).
\end{itemize}

We evaluate on \tbtwoone, 89 held-out tasks scored by execution. From a supervised
checkpoint at 49.4\%, three epochs of PPO on the quality-filtered \strategypool reach
\textbf{64.0\% resolved}, a \textbf{28.5\% relative} gain from RL alone. Under an identical
harness this places \Tone above GPT-5.4 at 54.8\% and DeepSeek-V4-Flash at 56.9\%,
approaching Claude Opus 4.7 at 66.1\%, the strongest model in its size band
(Section~\ref{sec:results}). Gains concentrate where terminal agency is tested:
\textbf{100.0} on debugging and \textbf{88.9} on system administration, both surpassing a
stronger general-purpose model.

Our training corpus is moreover fully \textbf{out-of-distribution} with respect to the
evaluation: isolated seeds and synthesized tasks disjoint from \tbtwoone ensures gains reflect genuine capability transfer over benchmark overfitting, so gains reflect transfer rather than benchmark
fitting (Section~\ref{sec:setup}).


\paragraph{Contributions.}
Our contributions are threefold:
\begin{itemize}[left=0pt]
    \item We introduce \Tone, a 122B MoE terminal agent trained purely by reinforcement
    learning on executed outcomes, up to \textbf{300+ tool-call turns} per task.

    \item We present a \textbf{stabilization stack} for large-scale sparse agentic RL,
    combining \TITO, \Rthree and a scheduled critic, with the infrastructure keeping a
    co-resident 122B actor--critic pair alive for days (Section~\ref{sec:infra}).

    \item We contribute a \textbf{dense execution reward} with its measured behaviour and
    two shaping variants that failed, alongside a candid record of failures
    (Section~\ref{sec:lessons}), which we found as instructive as the successes.
\end{itemize}

Together, these advances mark a significant step toward language models that act reliably
in consequential environments rather than merely describing how to do so.

\section{Training framework}
\label{sec:framework}

\begin{figure}[h]
  \centering
  \includegraphics[width=\textwidth]{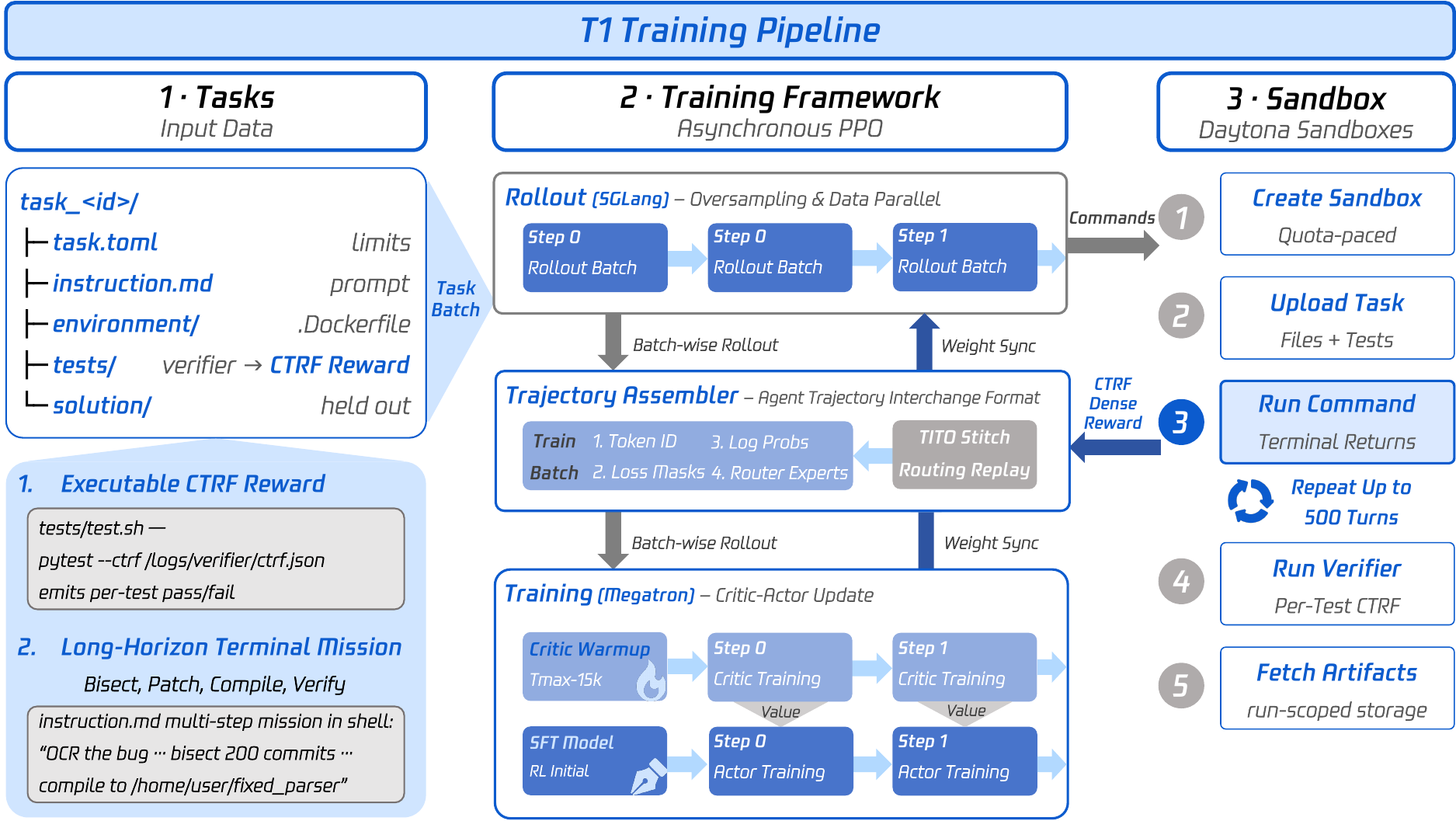}
  \caption{The \Tone training pipeline. \textbf{Left:} each task is self-contained,
  comprising metadata, a long-horizon instruction, an environment image with resource
  limits, a held-out verifier and a reference solution, and its verifier reports each
  assertion individually, which makes the reward executable rather than modelled.
  \textbf{Middle:} the inference replicas serve the behaviour policy under oversampling;
  the trajectory assembler normalizes interaction logs into \TITO-stitched training
  samples carrying routing records and the dense reward; the trainer updates the critic
  and then the actor before synchronizing weights back. \textbf{Right:} sandboxes are
  created, loaded, driven turn by turn, verified and reclaimed. Rollout and training run
  concurrently on disjoint devices.}
  \label{fig:arch}
\end{figure}

Figure~\ref{fig:arch} maps the system from left to right across three main components. A training backend and inference replicas run concurrently on disjoint accelerators under slime framework~\citep{slime}, pipelining step $t$ training with step $t+1$ generation to hide latency. Our terminal-agent integration attaches additively via public extension points. The following sections detail each ingredient: \textbf{Tasks} providing verifiable inputs, the \textbf{Training Framework} managing asynchronous rollouts and updates, and the \textbf{Sandbox} executing multi-turn commands for reward collection.

\subsection{Recursive Synthesis Terminal Training Tasks}
No reinforcement signal can be richer than what its verifier can measure, which makes task
construction a design decision rather than a preprocessing step;
Section~\ref{sec:setup} presents the resulting pools. Based on RST \citep{rst},
each task is self-contained: resource limits, a long-horizon instruction such as bisecting
hundreds of commits to locate a defect, patching it and proving the fix, a specification
from which an isolated container is built, a held-out verifier, and a reference solution
the agent never sees. The verifier is the load-bearing part, because it reports the
outcome of every assertion separately rather than a single pass or fail, which is what
makes the reward executed rather than modeled. \textbf{T1 utilized a selected proportion
of high-quality 15K from RST as training set }based on multi-dimensions of the tasks from
verifier, solution, instruction and value, following the audit developed in
Section~\ref{sec:data-construction}. A per-epoch seeded permutation then keeps the reward
distribution from drifting with task quality.

\subsection{Stable MoE Reinforcement Learning}
Between the trajectory a sandbox produces and the gradient the trainer applies lie dozens
of turns, two independent execution stacks and a discrete router, and each of them can
silently attribute the update to a policy that never generated the data. Inference
replicas serve the behaviour policy to many concurrent trials, and the scheduler
oversamples, admitting the first to complete and cancelling the straggler tail to bound
step time against heavy-tailed completions. Each trial drives its own sandbox, issuing
commands and reading their terminal output turn after turn until the task completes or its
limits are reached, whereupon the verifier runs and the sandbox is reclaimed, as
Section~\ref{sec:infra} describes. The agent exchanges token identifiers rather than text,
receiving back their log-probabilities and the expert routing chosen at every layer. An
assembler normalizes multi-turn logs into training samples in which sampled tokens carry
loss while tool output enters as masked context, and all four per-token streams are
transformed by the same offsets at every stage, which is the pipeline's central invariant.
Section~\ref{sec:stability} shows how three mechanisms build on it to close the gap above:
\textbf{Token-In-Token-Out}, \textbf{Routing Replay}, and \textbf{Infrastructure for Long-Horizon MoE Training}. The trainer updates the critic before the actor, since its pre-update values
anchor the advantage estimator and because both networks time-multiplex the same devices.
Weights are published once per step with generation quiesced first, so no request spans
two versions and the lag from pipelining is off-policyness of exactly one step.

\subsection{Dense Verification Reward Design}
A single rollout batch costs hundreds of sandbox-hours, and a binary outcome repays that
expense with one bit per trajectory; Section~\ref{sec:reward} spends the verifier's full
resolution instead. Each trial's per-assertion outcome becomes a \textbf{Dense Process
Reward}, scored by the absolute number of assertions satisfied on a scale fixed once for
the whole run, so that harder tasks carry proportionally more signal and the critic sees a
target comparable from step to step. The scalar enters at the final response token and the
critic distributes credit across the horizon; with a single sample per task there is no
group statistic to normalize against, leaving the critic as the only baseline. Because
such a reward can in principle be farmed rather than earned, we filter the pool for
verifiers too weak to validate their own goal and monitor trajectory growth throughout
training.

\subsection{Performance on Long-Horizon and Challenge Terminal Tasks}
To determine the overall performance of \Tone, we conduct comprehensive evaluation between multiple frontier models and baseline models \textbf{\tbtwoone \citep{terminalbench}, Long-Horizon Terminal-Bench \citep{li2026longhorizonterminalbenchtestinglimitsagents} and Terminal-Bench Hard \citep{rst}} in Section~\ref{sec:results}.
Three
epochs of PPO lift the supervised checkpoint from 49.4\% to \textbf{64.0\%} resolved on
\tbtwoone, placing \Tone above GPT-5.4 and DeepSeek-V4-Flash with an order of magnitude
fewer active parameters, and the gains hold where they matter most. On \lhtb, whose tasks
stress far longer horizons and are therefore the closer proxy for what this recipe
optimizes, \Tone reaches \textbf{27.9\%} and matches the performance of Gemini-3.1-Pro; on the harder Terminal-Bench Hard
subset it reaches \textbf{38.0\%}, ahead of DeepSeek-V4-Pro and well above both the
supervised and the base checkpoint.

\section{Terminal Dataset}
\label{sec:setup}

\subsection{Overview}
\label{sec:data}

A task is self-contained: per-trial limits, a long-horizon terminal instruction, an
environment specification, and a held-out verifier with a reference solution the agent
never sees. Three pools appear in our campaigns.

\begin{itemize}
\item \textbf{\tmax}: 14{,}601 tasks converted from the public corpus into
  terminal-bench layout. Its verifiers emit only a binary outcome with no per-assertion
  record, so only binary rewards are possible here.
\item \textbf{\rstpool}: 37{,}484 synthesized tasks generated via RST \citep{rst}, which iteratively extends seed solutions, realigns verifiers and instructions, and sandboxes each task before recursive seeding.
\item \textbf{\strategypool}: 15{,}000 tasks selected from the synthesis rounds by the audit of
  Section~\ref{sec:data-construction}. Their verifiers report per-assertion outcomes, and
  a pre-flight confirmed such records in 93\% of sampled tasks. The pool is materialized
  in quality-rank order, so per-epoch shuffling is mandatory
  (Section~\ref{sec:reward-data}).
\end{itemize}

\begin{figure}[h]
  \centering
  \includegraphics[width=1\linewidth]{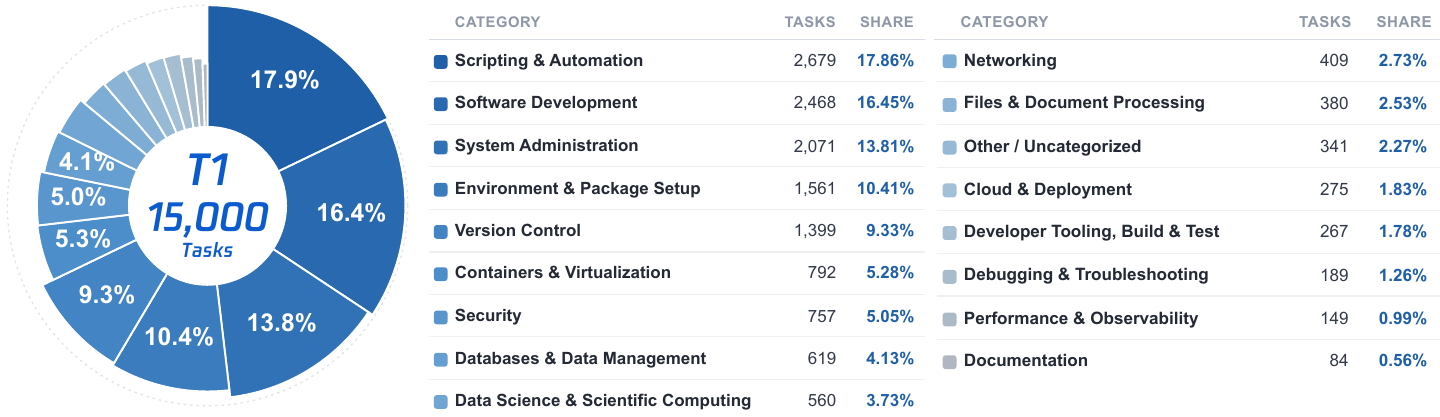}
  \vspace{0.1\intextsep}
  \caption{Category composition of \strategypool. All 15{,}000 tasks are counted once
  across 17 merged categories (from 47 raw labels); angle encodes share exactly, while
  radius is a rank-based power scale chosen to keep small slices visible and is therefore
  \emph{not} proportional to share. The pool is concentrated in command-line engineering
  work: the top five categories account for 67.9\% of all tasks.}
  \label{fig:data-categories}
\end{figure}

Because \strategypool carries the dense-reward runs, its composition is worth stating.
Figure~\ref{fig:data-categories} gives the breakdown: scripting and automation at
17.9\%, software development at 16.5\%, system administration at 13.8\%, environment and
package setup at 10.4\% and version control at 9.3\% together make up two-thirds of the
pool, while data science at 3.7\%, debugging at 1.3\% and performance work at 1.0\% are
thin. This skew predicts where residual failures land, as
Section~\ref{sec:case-study} confirms.

\subsection{Dataset Construction}
\label{sec:data-construction}

\begin{wrapfigure}{r}{0.49\linewidth}
  \centering
  \vspace{-0.6\intextsep}
  \includegraphics[width=\linewidth]{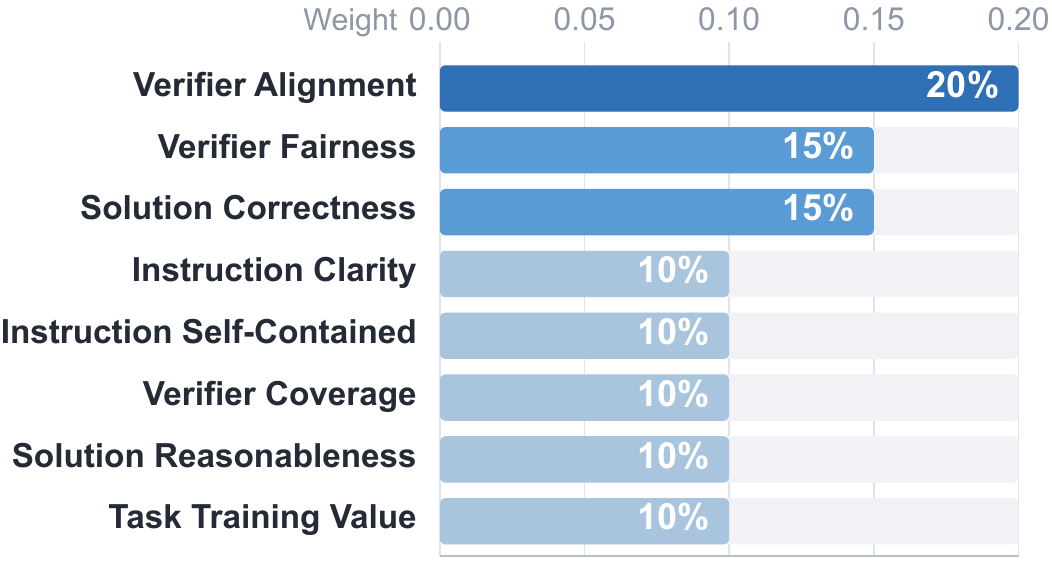}
  \caption{Figure 4: The eight audit dimensions and their weights, summing to 1.00. Aggregated by facet: Verifier $45\%$, Solution $25\%$, Instruction $20\%$, and Task Value $10\%$. Darker bars highlight the highest-weighted dimensions}
  \label{fig:data-scoring}
\end{wrapfigure}

Both synthesized pools are produced by recursive task synthesis \citep{rst}, which grows
a curriculum rather than sampling tasks independently. Each round takes accepted tasks
from the previous round as seeds under caps on parent lineage, category and rewrite
family; for each seed it selects a feasible rewrite operator, extends the reference
solution with additional executable steps, then aligns the environment, verifier and
public instruction to that longer workflow. Candidates are validated in a fresh sandbox
where the reference solution must genuinely pass the held-out verifier, with bounded
repair for recoverable failures and discard otherwise. Because difficulty is added to the
executable path before the instruction is rewritten, horizons lengthen without the task
degenerating into a longer prompt over the same behaviour.

Synthesis yield alone does not make a pool a usable RL signal, so \strategypool is the
subset of those rounds surviving an LLM audit. Figure~\ref{fig:data-scoring} shows what
that audit optimizes for. Alignment between the instruction and the verifier carries the
largest weight at $20\%$, because a mismatch between the public instruction and what the
held-out verifier enforces constitutes a hidden requirement and makes the task unusable:
the agent is punished for failing a criterion it was never shown. The remaining weight
splits between verifier quality, at $15\%$ for fairness and $10\%$ for coverage, and
solution quality, at $15\%$ for correctness and $10\%$ for reasonableness, with task
training value the only dimension exempt from the critical-minimum cutoff. Tasks are
hard-rejected for hidden requirements, test leakage, solution shortcuts, or verifiers too
weak to validate the goal. Selection ran in five stages over 15 rewrite rounds:
aggregation with lineage; static pre-check with executable validation; a semantic pass
yielding 5{,}902 accepted, 3{,}251 borderline and 5{,}847 rejected; instruction-only
repair of 6{,}875 tasks; and a re-audit.

\section{Stabilizing MoE RL training}
\label{sec:stability}

\subsection{Training-Inference Mismatch}
\label{sec:mismatch}

\paragraph{Setting.}
Rollout and training are served by two different systems: generation runs on inference
replicas built for throughput, with fused kernels, batched prefill and a paged key-value
cache, while the update runs on a training backend built for exact gradients, with its own
kernels, reduction orders and tensor layouts. The two agree on the parameters they hold
and on little else. Between them sits the agent harness, which exchanges no tensors at
all: it persists each assistant message as parsed text and re-renders the whole history
through a chat template before every turn.

The loop is additionally one-step asynchronous, so generation for step $t{+}1$ overlaps the
update at step $t$. Writing $\pi_t$ for the policy with parameters $\theta_t$, the batch
consumed by update $t$ was generated under $\pi_{t-1}$, and the per-token objective rests
on the ratio
\begin{equation}
  \label{eq:ppo-ratio}
  r_j(\theta) = \frac{\pi_{t}}{\pi_{t-1}} .
\end{equation}
That much is legitimate and fully modelled off-policyness of exactly one update, since
$\pi_{t-1}$ is a policy we did hold and the clip bounds how far $\pi_t$ may travel from it.
What is \emph{not} modelled is which system evaluates the denominator. We recompute it on
the training side, so validity demands that the trainer reproduce at version $t{-}1$ what
the sampler realized at that version. Marking evaluation by the sampler and by the trainer
with superscripts $\mathrm{r}$ and $\mathrm{t}$, the requirement
$\pi^{\mathrm{t}}_{t-1} \equiv \pi^{\mathrm{r}}_{t-1}$ separates into two independent
conditions.

\paragraph{Two fidelity conditions.}
Let $T_j$ be the token identifier at position $j$ of the assembled trajectory and
$\sI^{\ell}_j$ the \emph{routing mask} at MoE layer $\ell$, the set of experts the router
selects for that position by a discrete top-$k$ over $E$ candidates,
\begin{equation}
  \label{eq:moe}
  \sI^{\ell}_j = \operatorname{TopK}_k\!\big(\vs^{\ell}_j\big),
  \qquad
  \vy^{\ell}_j = \sum_{e \in \sI^{\ell}_j}
    \big[\vs^{\ell}_j\big]_e \, f_e\!\big(\vh^{\ell}_j\big),
\end{equation}
with $(L, k, E) = (48, 8, 256)$ for \ourmodel. Because $\sI_j = (\sI^1_j,\dots,\sI^L_j)$
decides \emph{which} experts act, it indexes a sub-network of the $95.5\%$ of parameters
held by experts, so the policy must be written $\pi_\theta(\cdot \mid T_{<j}, \sI_j)$. For
every position carrying loss, Equation~\ref{eq:ppo-ratio} therefore compares two versions
of one policy only if
\begin{align}
  \label{eq:tito-identity}
  \text{token fidelity} \qquad
  & T^{\mathrm{t}}_j = T^{\mathrm{r}}_j , \\
  \label{eq:r3-identity}
  \text{routing fidelity} \qquad
  & \sI^{\mathrm{t},\ell}_j = \sI^{\mathrm{r},\ell}_j
    \quad \forall\, \ell \in [L] .
\end{align}
The harness breaks token fidelity: re-rendering the history returns turn $i$'s output as
$\mathrm{enc}(\mathrm{dec}(\va_i))$, and that round trip is not the identity whenever
parsing normalizes the message or the template re-tokenizes at a boundary, so the trainer
conditions on a stream the sampler never produced. The two stacks break routing fidelity:
they compute $\vs^{\ell}_j$ by different kernels, and since $\operatorname{TopK}_k$ is
discontinuous, a numeric difference far below any tolerance one would place on a logit
suffices to exchange a selected expert for its runner-up, substituting one sub-network for
another so that the ratio relates two different networks rather than two versions of one.

\paragraph{Why both must be enforced.}
The conditions are independent, so enforcing either leaves the other's failure mode intact.
Dense models cannot violate routing fidelity at all and short-horizon tasks make token
fidelity nearly automatic, whereas the model emitting tens of tool-calling turns violates
both across roughly $10^4$ loss-bearing positions, where per-position discrepancies
accumulate along the trajectory instead of cancelling. Section~\ref{sec:tito} enforces
token fidelity by having the trainer consume the identifiers the sampler emitted, repairing
turn boundaries under a small auditable set of cases; Section~\ref{sec:r3} enforces routing
fidelity by recording $\sI^{\mathrm{r},\ell}_j$ during generation and replaying it in the
training forward pass. Appendix~\ref{app:notation} tabulates every symbol, and
Appendix~\ref{app:async} separates the discrepancy attributable to version skew, which
should be nonzero, from the cross-system component these mechanisms remove.

\subsection{\TITO: token-in, token-out}
\label{sec:tito}

Writing $\mT_i = \vp_i \Vert \va_i$ for the stream turn $i$ contributes, \TITO asks that
$\mT_i$ be a \emph{bit-exact prefix} of $\vp_{i+1}$ at every boundary: one displaced
identifier replaces $\pi(T_j \mid T_{<j})$ by $\pi(T_j \mid \tilde{T}_{<j})$ at that
position and every position after it. Three harness behaviours break the requirement.
Encoding is canonical while decoding is many-to-one, so a non-canonical sampled split is
lost to its canonical re-encoding; templates prune reasoning before the last
\textsf{User} message, which the harness emits once per observation; and a re-serialized
tool call returns different whitespace, hence different identifiers.

\paragraph{Token-in preserves prefixes to prevent re-encoding drift.}
With $\gM_i$ the message history, $\vp_i = \mathrm{enc}(\mathrm{template}(\gM_i))$ and
$(\va_i, \vq_i) = \textsc{Sample}_{\pi^{\mathrm{r}}_{t-1}}(\vp_i)$ comes from the sampler's
per-token output, so $\mathrm{enc}$ acts once per turn rather than once per history replay,
with templates pinned to keep rendering append-only.

\paragraph{Token-out stitches streams under loss masking.}
A trial becomes one stream $\mT = (T_1,\dots,T_N)$ with mask $\vm$ and log-probabilities
$\vq$ satisfying
\begin{equation}
  \label{eq:mask}
  m_j = 1 \iff T_j \text{ was sampled by } \pi^{\mathrm{r}}_{t-1},
  \qquad
  q_j = m_j \cdot \log \pi^{\mathrm{r}}_{t-1}\big(T_j \mid T_{<j}, \sI^{\mathrm{r}}_j\big),
\end{equation}
so observations and glue give context but no gradient. Between turns the assembler tests
progressively weaker prefix relations between $\mT_i$ and $\vp_{i+1}$, and the case it lands
in, enumerated in Box~\ref{box:tito-cases}, determines how the boundary is repaired.

\begin{casebox}{\textbf{TITO Hierarchy.} Boundary cases evaluated at each turn from strongest to weakest.}{box:tito-cases}
Each line pairs the condition under which turn $i$ may be appended with what the assembler
then appends, the first admissible one deciding the boundary.
\begin{align}
  \label{eq:case-strict}
  \textsc{strict}\;&:\; \mT_i \preceq \vp_{i+1},
    \quad \text{context} = \vp_{i+1} \ominus \mT_i; \\
  \label{eq:case-norm}
  \textsc{normalized}\;&:\; \textstyle\min_{(s,u)}(s{+}u) \text{ s.t. }
    \mathrm{drop}_s(\vp_i) \Vert \mathrm{drop}_u(\va_i) \preceq \vp_{i+1},
    \;\, (s,u) \in [0,96] \times [0,16]; \\
  \label{eq:case-retok}
  \textsc{retokenized}\;&:\; \mathrm{dec}(\va_i) = \mathrm{dec}(\tilde{\va}_i),
    \;\, \tilde{\va}_i = \vp_{i+1}[b\!:\!e] \text{ by offset mapping}; \\
  \label{eq:case-split}
  \textsc{split}\;&:\; \text{neither holds, so a new chunk opens under the same trial.}
\end{align}
\end{casebox}

Case~\ref{eq:case-strict} is exact \TITO; Case~\ref{eq:case-norm} is a bounded repair on a
finite $97 \times 17$ grid whose $96$ is the shared template suffix, which keeps it
auditable. Case~\ref{eq:case-retok} would break Equation~\ref{eq:tito-identity} wholesale,
so the assembler falls back to text-space equality and appends
\begin{equation}
  \label{eq:retok-substitution}
  \mT \mathrel{+}= \underbrace{\va_i}_{m = 1}
     \Vert \underbrace{\vp_{i+1}[e\!:\!]}_{m = 0},
  \qquad \text{never } \tilde{\va}_i .
\end{equation}

\begin{figure}[t]
  \centering
  \includegraphics[width=\linewidth]{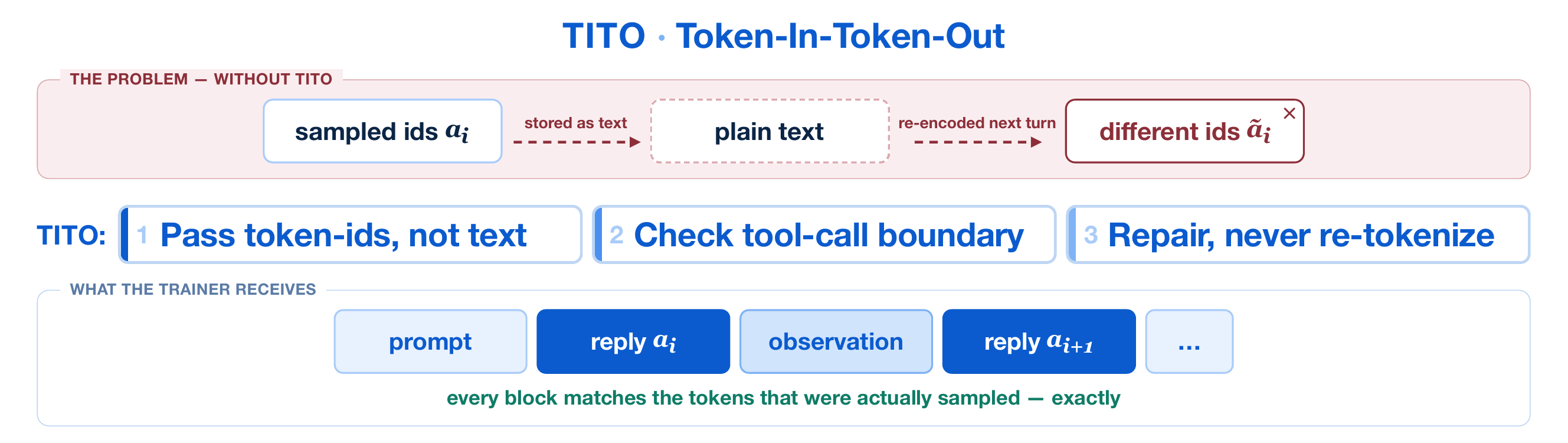}
  \caption{\TITO stitching, drawn for the \textsc{retokenized} case
  (Equation~\ref{eq:case-retok}), the hardest of the four. The re-tokenized copy
  $\tilde{\va}_i$ never enters the training stream, whereas the sampled $\va_i$ does;
  observations and glue enter masked, with $m{=}0$ and $q{=}0$.}
  \label{fig:tito}
\end{figure}

\paragraph{Empirical verification confirms zero drift.}
\label{sec:tito-audit}
An auditor re-locates $\va_i$ inside $\vp_{i+1}$ per transition and checks span and text
equality, the right test since Case~\ref{eq:case-retok} makes token equality flag harmless
re-encodings. Splitting the $N$ positions into aligned $\gA$, re-tokenized $\gD$ and
placeholder $\gP$, a production run over 1{,}402 samples puts the drift rate inside the
loss region at \textbf{exactly zero}:
\begin{equation}
  \label{eq:audit}
  \frac{|\{j \in \gD \cup \gP : m_j = 1\}|}{|\{j : m_j = 1\}|} = \mathbf{0.0000\%},
\end{equation}
so every position that carries gradient satisfies Equation~\ref{eq:tito-identity} exactly.

\subsection{\Rthree: rollout routing replay}
\label{sec:r3}

\Rthree \citep{r3} enforces Equation~\ref{eq:r3-identity} by construction: record the
routing mask inference selected, then reuse it in the training forward pass. The primitives
are available upstream; we supply the capture path, the multi-turn alignment and the
failure policy.

\paragraph{Rollout routing replay.}
Conventionally the training pass derives both quantities of Equation~\ref{eq:moe} from its
own logits, selecting $\operatorname{TopK}_k(\vs^{\ell}_j(\theta_t))$ and normalizing over
that selection. \Rthree keeps the normalization on the live logits but takes the selection
from the recorded mask $\sI^{\mathrm{r},\ell}_j$, renormalizing over the recorded experts
alone,
\begin{equation}
  \label{eq:r3-inject}
  g^{\ell}_{j,e}
  = \frac{\exp\big(\big[\vs^{\ell}_j(\theta_t)\big]_e\big)}
         {\sum_{e' \in \sI^{\mathrm{r},\ell}_j}
          \exp\big(\big[\vs^{\ell}_j(\theta_t)\big]_{e'}\big)}
  \;\; \text{for } e \in \sI^{\mathrm{r},\ell}_j,
  \qquad
  \vy^{\ell}_j = \sum_{e \in \sI^{\mathrm{r},\ell}_j}
    g^{\ell}_{j,e}\, f_e\!\big(\vh^{\ell}_j\big).
\end{equation}
This serves two purposes. It \emph{aligns} training with inference, since the experts
carrying gradient are exactly those that produced the sample, removing the discontinuity
that made an exchanged expert possible. It also \emph{preserves the gradient path}: only
the mask is replayed while the softmax still acts on $\vs^{\ell}_j(\theta_t)$, leaving the
router trainable and the computation graph untouched. The trainer's router is wrapped, not
reimplemented. Replay is only as good as the record, however, and Box~\ref{box:r3-cost}
states what keeping one costs.

\begin{casebox}{\textbf{Cost Analysis.} Negligible rollout overhead with zero additional arithmetic.}{box:r3-cost}
The sampler returns a compact integer tensor $\tR_i \in [E]^{\rho_i \times L \times k}$ per
turn, one row per predicting position:
\begin{equation}
  \label{eq:rows}
  \tR_i[j, \ell, :] = \sI^{\mathrm{r},\ell}_j,
  \qquad
  \rho_i = |\mT_i| - 1,
  \qquad
  b_{\text{tok}} = L k \cdot 4\,\text{B} = \mathbf{1536\,\text{B/token}},
\end{equation}
the $-1$ because the final token predicts nothing. At $(L,k) = (48,8)$ this is $1.5$\,KiB
per position, or some $48$\,MiB for a $33$k-token trajectory, and it holds the rollout
overhead below $3\%$: the masks are a by-product of a forward pass that already computed
them, so capture adds transport but no arithmetic.
\end{casebox}

\paragraph{Mask caching and multi-turn alignment.}
Recorded masks inherit the property that makes prefix caching sound: for identical prefix
tokens the router yields identical selections, so masks are cached alongside the key-value
cache and reused on a prefix hit. This matters because every tool call resumes a shared
prefix, and re-prefilling purely to regenerate masks would dominate a long trajectory. On
the training side the records ride the same stitching as tokens, keeping
$\tR[j,\ell,:] = \sI^{\mathrm{r},\ell}_j$ at the index $j$ that also indexes
$(T_j, m_j, q_j)$ in Equation~\ref{eq:mask}. At repaired boundaries a record may belong to
no turn's capture, and a neighbour is substituted only where no gradient is touched:
\begin{equation}
  \label{eq:placeholder}
  \tR[j,:,:] \gets \tR[j{-}1,:,:]
  \quad\text{iff}\quad m_{j+1} = 0,
  \qquad\text{otherwise raise},
\end{equation}
so replayed routing is approximate only on the set Equation~\ref{eq:audit} already excludes
from the loss. Absent routing is a hard error and the rollout aborts rather than dropping
the sample, which would condition the batch on capture having succeeded. Records also
receive exactly the token pipeline's sharding, since any other composition would silently
replay the wrong experts, which is why $\rho = N - 1$ is asserted per sample;
Appendix~\ref{app:align} gives the invariant chain.

Algorithm~\ref{alg:r3} gives the schedule, with two consequences. The reference pass selects
freely, so $\KL(\pi_\theta \| \pi_{\mathrm{ref}})$ is positive at step $0$ by design and any
assertion of a vanishing initial divergence must be disabled. Separate forward and backward
cursors are needed, since activation recomputation re-runs each layer's forward pass during
the backward pass. The critic never replays, its target requiring no behavioral fidelity to
the sampler.

\begin{algobox}{Routing-replay schedule for one training step. The router operates in
three modes: free selection, replay without gradient, and replay with gradient.}{alg:r3}
\Require per-turn captures $\{(\vp_i, \va_i, \vq_i, \tR_i)\}_{i=1}^{T}$ from the sampler
\For{each turn $i$}
  \State $(\va_i, \vq_i) \gets \textsc{Sample}_{\pi^{\mathrm{r}}_{t-1}}(\vp_i)$ \textbf{with}
         $\tR_i[j,\ell,:] \gets \sI^{\mathrm{r},\ell}_j$
         \Comment{Equation~\ref{eq:rows}}
\EndFor
\State $(\mT, \vm, \vq) \gets \textsc{Stitch}(\cdot)$ under
       cases~\ref{eq:case-strict}--\ref{eq:case-split};\;
       substitute a record only where $m_{j+1} = 0$
       \Comment{Equation~\ref{eq:placeholder}}
\State \textbf{abort} if any capture is missing;\;
       \textbf{assert} $\rho = |\mT| - 1$;\; shard $\tR$ as $\mT$
\State \textbf{free:} evaluate $\log \pi_{\theta_{\mathrm{ref}}}(\mT)$
       \Comment{$\KL > 0$ at step $0$}
\State \textbf{replay, with gradient:}
       $\theta \gets \theta - \eta_\theta \nabla_\theta \gL^{\text{PPO}}$ with
       $r_j(\theta)$ from Equation~\ref{eq:ppo-ratio}
\State \textbf{free:} $\phi \gets \phi - \eta_\phi \nabla_\phi \gL^{V}$;\;
       release the record
\end{algobox}

\subsection{Measuring Training Stability of T1}
\label{sec:mismatch-effect}
\label{sec:other-stability}

\begin{figure}[t]
  \centering
  \begin{minipage}[c]{0.615\linewidth}
    \centering
    \includegraphics[width=\linewidth]{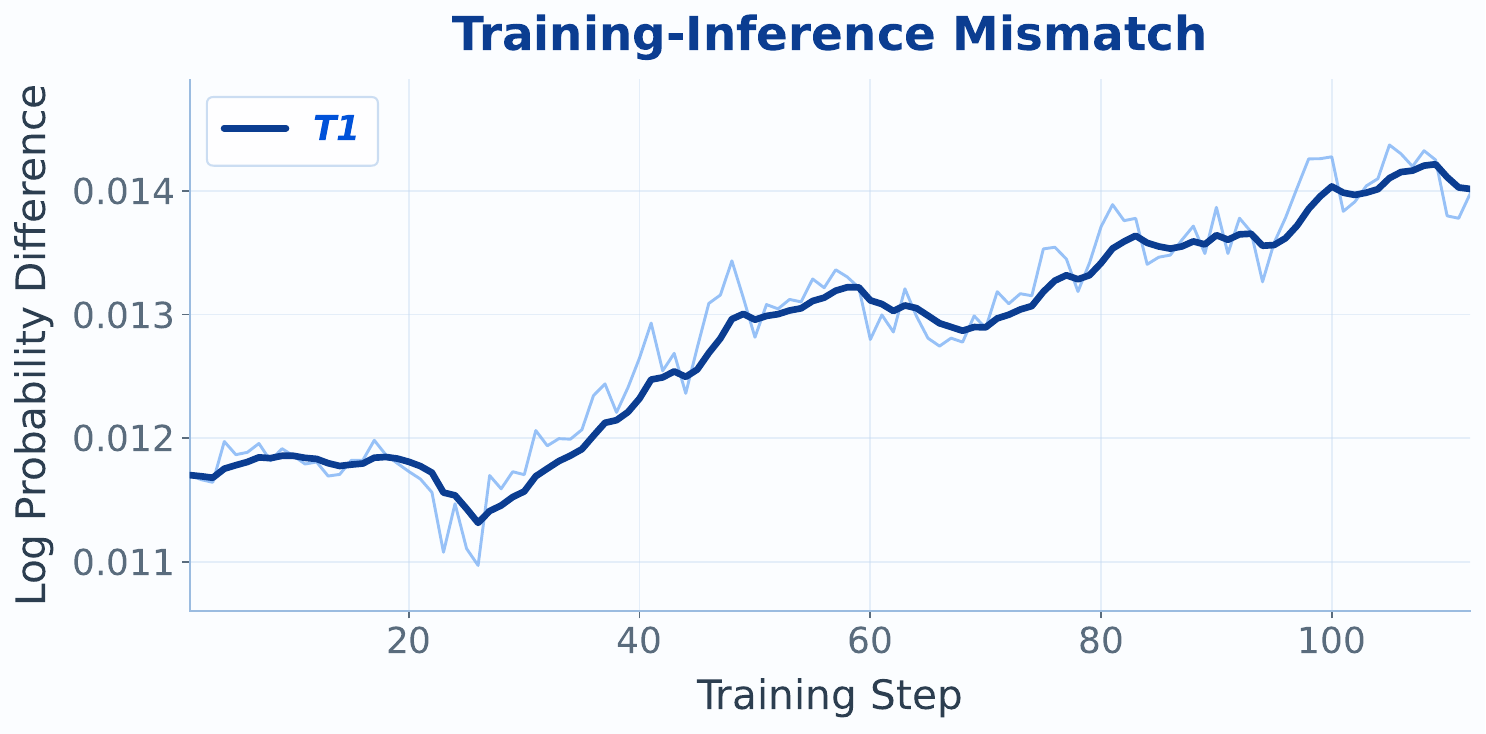}
  \end{minipage}\hfill
  \begin{minipage}[c]{0.355\linewidth}
    \centering
    \tablesize
    \begin{tabular}{@{}lc@{}}
    \toprule
    Configuration & $|\Delta \log p|$ \\
    \midrule
    \rowcolor{metabg!8} \Tone (\TITO $+$ \Rthree) & \textbf{0.013} \\
    without \TITO and \Rthree & 0.021 \\
    \bottomrule
    \end{tabular}
  \end{minipage}
  \caption{Train--inference log-probability gap $|\Delta \log p|$
  (Equation~\ref{eq:tis}). \figleft{} Per-step series over the production dense-reward run,
  with an exponential moving average at $\alpha{=}0.25$ in bold. \figright{} Mean gap on the
  same 122B stack with and without the two mechanisms. 
  }
  \label{fig:logprob-diff}
\end{figure}

\paragraph{Measure the Training-Inference Mismatch.}
Following \citet{sat}, the realized mismatch at a loss-bearing position is the log of the
sampled importance ratio:
\begin{equation}
  \label{eq:tis}
  d_j \;=\; \log \pi_{t} - \log \pi_{t-1}.
\end{equation}
At a staleness of one update the first term is the single optimizer step the clip is
designed to correct, and it should be nonzero; the second is nonzero even at identical
weights, because the two stacks differ in kernels, reduction orders and routing, and it is
precisely what \TITO and \Rthree remove. As shown in Figure \ref{fig:logprob-diff}, we report the mask-weighted mean of $|d_j|$ over
the loss region, computed outside the autograd path so the recipe is bit-identical whether
or not it is collected. The reduction matters: normalizing by micro-batch count rather than
by $\sum_j m_j$ inflates the statistic by three orders of magnitude at unit micro-batch.

\paragraph{\Tone training stability.}
\label{sec:stability-indicators}
Each mechanism pairs with a quantity that certifies it. Token fidelity is confirmed by
Equation~\ref{eq:audit}, which puts drift inside $\{j : m_j = 1\}$ at $0.0000\%$. Routing
fidelity is enforced by substitution rather than measured, since
Equation~\ref{eq:r3-inject} replaces $\operatorname{TopK}_k$ by a lookup, leaving only the
placeholder set of Equation~\ref{eq:placeholder}, bounded at $0.003\%$ and disjoint from
the loss region. Jointly they move the gap from $0.021$ to $0.013$, which decides whether
$r_j(\theta)$ reflects policy movement or bookkeeping error. The residual is expected, as
\Rthree aligns expert selection but not kernel numerics, and a slow upward drift is no
regression either, since $\Delta^{\pi}_j$ grows when the policy legitimately improves.

\paragraph{Algorithmic measures that keep the run stable and efficient.}
We keep the objective deliberately spare and bound the cost of a single step, withholding every optional term that fought an alignment mechanism or added a gradient the reward does not justify:
\begin{itemize}
    \item \textbf{Surrogate clipping \& KL penalties:} The surrogate clips symmetrically at \textbf{$\varepsilon = 0.2$}, while both KL terms are disabled because the frozen reference routes with its own selection and would unfairly charge the policy for a bookkeeping difference.
    \item \textbf{MoE load balancing:} The load-balancing coefficient is set to \textbf{zero}, as balancing pressure asks the router to redistribute exactly the choices replay asks it to reproduce.
    \item \textbf{Optimizer configuration:} Optimization is performed using Adam with \textbf{$\beta = (0.9, 0.98)$}, weight decay of $0.1$, and a constant learning rate.
    \item \textbf{Oversampling:} Long-horizon trials have a heavy length tail, so a step that waits for every trajectory is paced by its slowest few. We therefore train at a batch of \textbf{560} and oversample during rollout, admitting the first \textbf{512} trajectories to complete and utilize data parallel at \textbf{8} to accelerate training, truncating the remaining tail, which bounds step time at the price of a mild bias against the longest trials.
    \item \textbf{Compilation storms:} A compilation storm is indistinguishable from a collective hang, one rank having once compiled for over $30$ minutes while its peers waited inside the all-to-all.
\end{itemize}
The scheduling and liveness machinery that realizes the last two items is described in Section~\ref{sec:infra}.

\subsection{Critic-side stability}
\label{sec:critic}

PPO here uses a separate, full-size critic: a second copy of the same architecture whose
language-model head is replaced by a scalar value head on the last pipeline stage. It
shares the actor's device allocation without additional hardware, and training-side
offload is forced so that the two 122B networks time-multiplex those devices. Each must
therefore fit alone in 95\,GiB, which is the constraint driving
Section~\ref{sec:infra-memory}.

\paragraph{Ordering.}
The critic trains first each step and hands its pre-update values
$V_{\mathrm{old}} = V_{\phi_{t-1}}$ to the actor, anchoring both the advantage estimator
and the value objective to one fixed function:
\begin{gather}
  \label{eq:ppo_updates}
  \hat A_j = \sum_{n \ge 0} (\gamma\lambda)^{n} \delta_{j+n}, 
  \quad \text{where} \quad 
  \delta_j = \hat r_j + \gamma V_{\mathrm{old}}(s_{j+1}) - V_{\mathrm{old}}(s_j), \\
  \mathcal{L}^{V}(\phi) = \mathbb{E}_j \Big[ \max\big(
      (V_\phi(s_j) - \hat R_j)^2,
      (V_{\mathrm{clip}}(s_j) - \hat R_j)^2
  \big) \Big], \\
  \mathcal{L}^{\pi}(\theta) = -\mathbb{E}_j \Big[ \min\big(
      r_j(\theta) \hat A_j,
      \operatorname{clip}(r_j(\theta), 1 - \epsilon, 1 + \epsilon) \hat A_j
  \big) \Big],
\end{gather}
where $V_{\mathrm{clip}}(s_j) = V_{\mathrm{old}}(s_j) + \operatorname{clip}(V_\phi(s_j) - V_{\mathrm{old}}(s_j), -\epsilon_v, \epsilon_v)$ 
and $r_j(\theta) = \frac{\pi_\theta(a_j \mid s_j)}{\pi_{\theta_{\mathrm{old}}}(a_j \mid s_j)}$ 
denotes the probability ratio. Here, $\gamma = \lambda = 1$, and the clipping thresholds 
are set to $\epsilon = 0.2$ for the actor policy and $\epsilon_v = 0.2$ for the critic value function.

\paragraph{Critic Warm-Up.}
Critic Warm-Up trains the value network as well as the actor, with only critic model being saved, for one epoch over
\tmax before any policy step is taken. The dense-reward campaign loads those weights,
weights only so that no optimizer moment crosses runs, and needs just $N{=}2$
re-calibration rollouts, whereas the binary-reward campaign cold-started its critic
instead.

\begin{figure}[t]
  \centering
  \includegraphics[width=0.7\linewidth]{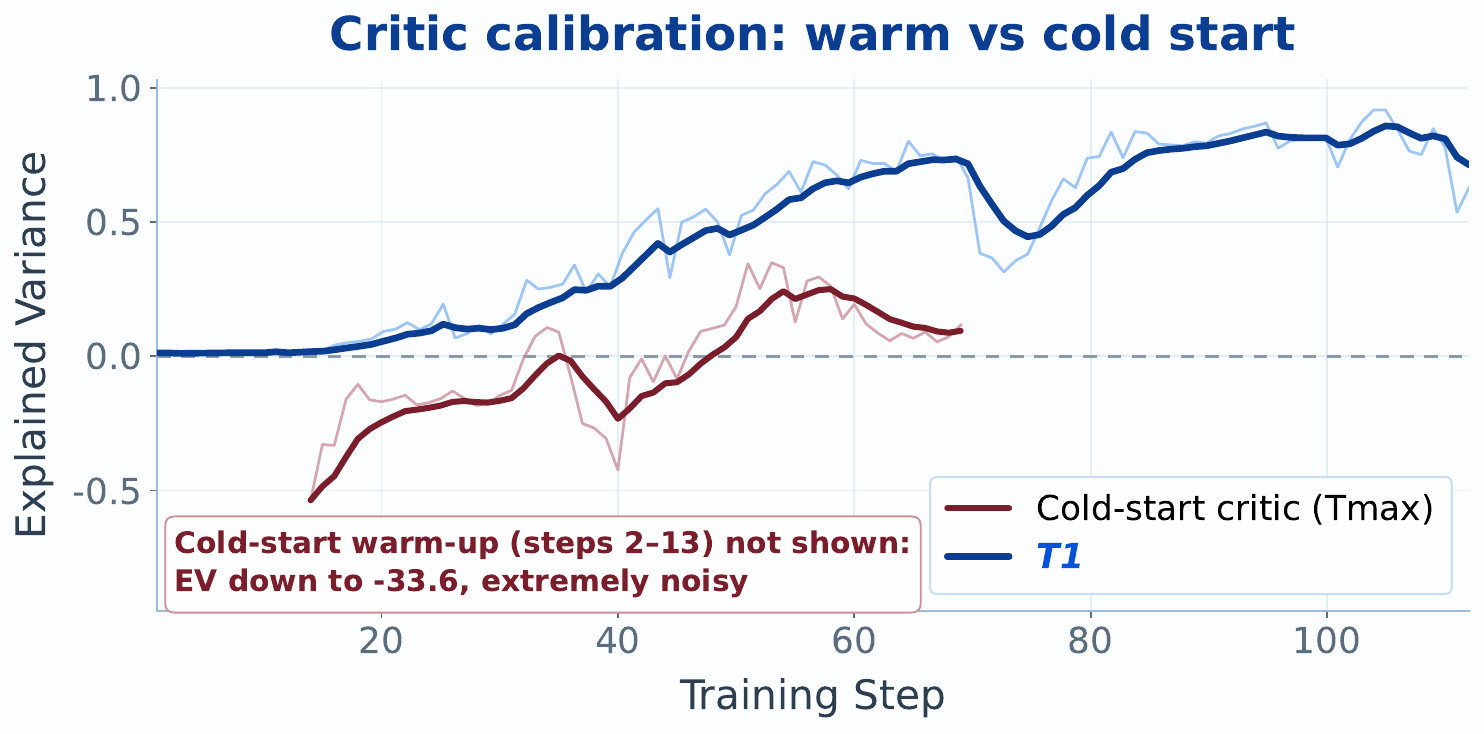}
  \caption{Critic explained variance, defined in Equation~\ref{eq:ev}, with and without
  Critic Warm-Up. \textbf{Blue (\Tone)}: the production dense-reward run on \strategypool,
  whose critic comes from Critic Warm-Up over \tmax. \textbf{Red}: the cold-started
  critic of the \tmax campaign. Faint lines are per-step values and bold lines an
  exponential moving average at $\alpha{=}0.25$, while the dashed rule marks
  $\mathrm{EV}{=}0$. The cold start starts at $\mathrm{EV}{=}-33.6$ and is negative for 30 of 58 logged steps, whereas the Critic Warm-Up run never
  goes negative and plateaus between $0.71$ and $0.86$.}
  \label{fig:critic-ev}
\end{figure}

What this buys is read off explained variance, the fraction of return variance the value
function accounts for, computed over the tokens of a rollout batch with GAE returns
$\hat R_j$:
\begin{equation}
  \label{eq:ev}
  \mathrm{EV} = 1 - \frac{\Var_j\!\big[\hat{R}_j - V_\phi(s_j)\big]}
                        {\Var_j\!\big[\hat{R}_j\big]} ,
\end{equation}
where both variances are reduced globally across context- and data-parallel ranks, since a
per-rank value averaged across ranks is biased. The quantity is unbounded below, and
$\mathrm{EV}<0$ means that subtracting $V_\phi$ adds variance to $\hat A_j$ instead of
removing it. A cold start opens at $-33.6$ and spends roughly the first half of the
campaign paying down that deficit, whereas Critic Warm-Up settles between $0.71$ and
$0.86$ from the first update. The same measurement fixes the learning rates: the critic
runs at $1.5\times10^{-5}$ against $1.0\times10^{-6}$ for the actor, absorbing the larger
step because its target is a supervised regression rather than a policy improvement, and on
27B pathfinding runs moving this ratio from $10\times$ to $20\times$ lifted $\mathrm{EV}$
from $-39$ to $+0.11$ while $30\times$ shortened Critic Warm-Up further. Reward stayed flat
across these settings, which redirected attention to the data as
Section~\ref{sec:reward} describes.

\paragraph{Value-target conditioning.}
Two choices keep the regression target well-posed. The dense reward uses a global fixed
scale, as Section~\ref{sec:reward-def} sets out, and the trajectory reward lands on the
final token with $\gamma = \lambda = 1$, which makes per-token returns piecewise constant
so that the value problem reduces to predicting a trajectory's final score from its
prefix. 

\section{Dense Verification Reward Design }
\label{sec:reward}

\subsection{From sparse outcomes to dense verification signals}
When we began RL on \strategypool, many tasks were too difficult for the model to solve completely. Under a binary task-solved reward, these unsuccessful trajectories all received zero, even when the agent had satisfied some of the task's requirements. Complete successes were too rare to provide a useful learning signal, while partial progress remained invisible to the reward. Learning on these tasks therefore required feedback that could distinguish degrees of completion before the model could reliably produce a full solution. 

Such feedback depends on the verifiers supplied with the training data. In our earlier RST work, we anticipated this need during task synthesis and equipped the synthesized tasks with sufficiently many verification checks covering individual task requirements. These checks make partial completion observable through per-assertion outcomes. By comparison, \tmax lacks a sufficiently rich set of verification checks to support this form of dense reward. The RST synthesis process thus provides the foundation for dense feedback in our synthesized training pools, including \strategypool. 

We use these verification outcomes to reward the number of assertions an agent satisfies, giving credit for partial solutions even when the overall task remains unsolved. This turns otherwise zero-reward trajectories into graded supervision and allows the model to learn from progress on tasks it cannot yet complete. The following subsection defines how these per-assertion outcomes are converted into a scalar reward.

\subsection{Per-assertion verification and the test-count reward}
\label{sec:reward-def}

Each task's verifier is invoked so that it emits a structured per-assertion report inside
the sandbox. After rollout the reward stage reads that report back from the trial and
computes, with $P$ the number of passing assertions,
\begin{equation}
  \label{eq:reward-def}
  r \;=\; \frac{P}{S}, \qquad S = 20 ,
\end{equation}
that is the absolute passing count on a fixed global scale, and explicitly not the pass
ratio. Three decisions are worth stating.

\begin{itemize}
\item \textbf{Absolute count rather than ratio.} Our training pool mixes tasks of
  different difficulty, and we shuffle them together without an easy-to-hard curriculum.
  A single batch therefore contains both easy and hard tasks, making the reward scale
  across tasks consequential. For example, passing 10 of 20 assertions on a hard task
  and passing 2 of 4 on an easy task both yield a pass ratio of $0.5$. Yet satisfying
  those ten assertions can require substantially more work, potentially through a long
  sequence of tool interactions. The equal ratios hide this difference in verified
  progress.
  Our synthesis process was designed with this comparison in mind: harder tasks were
  equipped with more verification checks, while each check was intended to represent
  a roughly comparable increment of work across tasks. This is an approximate design
  principle, rather than a guarantee that all assertions require identical effort.
  Under this principle, the absolute passing count better reflects the amount of
  verified progress in a mixed batch. With the shared scale $S=20$, the two trajectories
  above receive $10/20=0.5$ and $2/20=0.1$, respectively. Each additional passing
  assertion contributes the same $1/S$ reward, preserving the intended distinction
  between completing more requirements on a hard task and fewer on an easy one.
\item \textbf{Fixed global scale.}  We chose $S=20$ after measuring the assertion-count
  distribution across the 15{,}000 tasks in \strategypool. This value sits near the
  ninetieth percentile of that distribution, whose median is 4 and maximum roughly 35.
  We keep it fixed throughout training: a per-batch maximum would make the reward
  scale drift from step to step and hand the
  critic an inconsistent regression target, whereas cross-step consistency is precisely
  what makes the value function learnable.
\item \textbf{Fallback.} If the per-assertion report is missing or unparsable the trial
  falls back to the binary terminal outcome, so a genuinely solved task never scores
  zero, and parse failures are tagged by cause for observability.
\end{itemize}

An earlier ratio-based variant, $r=\max(b,\,0.4\,P/T)$ clamped to $[0, \max(1,b)]$ with
$T$ the total assertion count and $b$ the terminal outcome, is retained for controlled
comparison. Algorithm~\ref{alg:reward} states the computation as used in production.

\begin{figure}[t]
  \centering
  \includegraphics[width=0.65\linewidth]{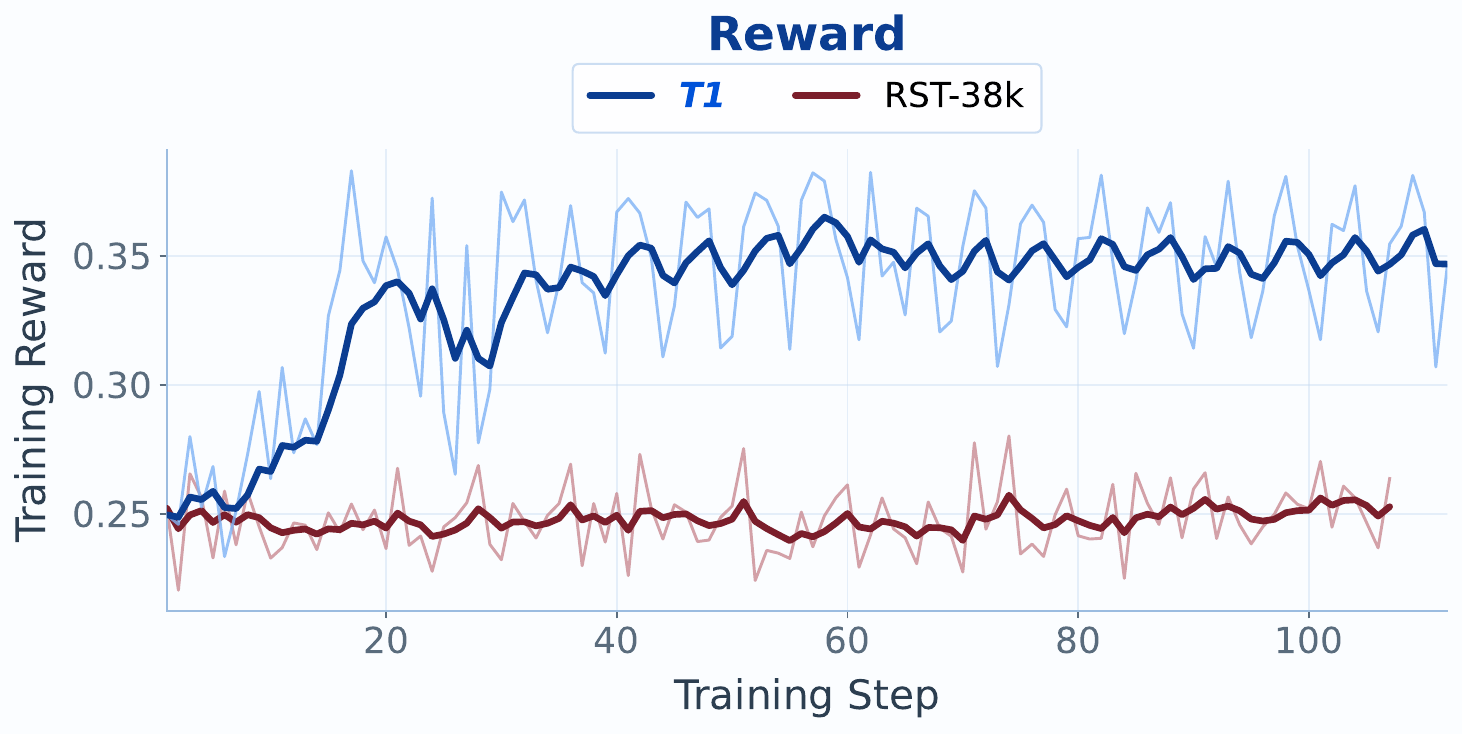}
  \caption{Mean rollout reward under the test-count reward over the production
  dense-reward run, reported before any normalization, since under PPO the reward stage
  returns raw values as Algorithm~\ref{alg:reward} states. Faint line: per-step mean;
  bold line: exponential moving average at $\alpha{=}0.25$. Reward rises from 0.250 to
  roughly 0.345 in the first 50 steps and then holds a band of 0.34 to 0.36 for the
  remaining 60.}
  \label{fig:reward-curve}
\end{figure}

Figure~\ref{fig:reward-curve} shows the reward this definition actually produces in
training, and it is worth reading against the design choices above. The absolute scale
behaves as intended: a mean near 0.35 at $S{=}20$ corresponds to roughly 7 passing
assertions per trajectory, comfortably inside the resolution of the signal rather than
pinned at either end, which is precisely what the ratio formulation would have destroyed
by capping every task at $1.0$. The trajectory has two phases, a steep climb from 0.250
to roughly 0.345 over the first 50 steps and then a plateau in the band 0.34 to 0.36 for
the remaining 60 steps, peaking at 0.365 near step 58. The plateau is not stagnation,
since the held-out benchmark keeps improving through it as Section~\ref{sec:dynamics}
shows, so the run redistributes which assertions it satisfies rather than simply
satisfying more of them. This is consistent with per-epoch reshuffling keeping the
per-batch reward statistics stationary rather than letting them drift with task quality.
The visible high-frequency oscillation is the batch-composition signature of that
shuffling at batch 512 with one sample per task rather than an instability: its amplitude
of about 0.02 stays constant over the run, and the band never collapses toward zero.

\begin{algobox}{Dense reward assignment for one rollout step. $P$ is the number of
passing assertions reported by the verifier, $b \in \{0,1\}$ the terminal outcome, and
$S$ the fixed normalizer.}{alg:reward}
\Require completed trials of the current step, grouped by task and trajectory
\For{each trial}
  \State $b \gets$ terminal verifier outcome
  \If{a per-assertion record is available}
    \State $r \gets P / S$ 
    \Comment{$S = 20$; $r$ may exceed $1$}
  \Else
    \State $r \gets b$ \Comment{a solved task is never assigned zero}
  \EndIf
  \State broadcast $r$ to every chunk of the trial, so credit is trajectory-level
\EndFor
\State return $r$ unnormalized, since with one sample per task no group baseline exists
\State place $r$ on the final response token; GAE with $\gamma = \lambda = 1$ against the
       critic distributes credit over the horizon
\end{algobox}

\subsection{Credit assignment}
Density in Equation~\ref{eq:reward-def} concerns the reward's value resolution rather than
its temporal placement: the score gains $1/S$ per additional satisfied assertion, yet is
still delivered as one scalar on the last response token of each trajectory chunk. This formulation has two direct implications for learning:

\begin{itemize}
\item \textbf{Temporal credit comes from the value function.} GAE at
  $\gamma = \lambda = 1$ propagates the terminal scalar backward over the interaction
  horizon, and no potential-based shaping or per-turn term is added.
\item \textbf{The reward stays unnormalized.} Each task contributes one trajectory per
  step, so no group statistic exists from which a group-relative baseline could be formed,
  and advantage normalization would rescale away the cross-task differences an absolute
  passing count is meant to preserve.
\end{itemize}

The critic is thus the only baseline, and its calibration in Section~\ref{sec:critic}
therefore lies on the critical path.

\subsection{Preventing reward hacking}
\label{sec:reward-antihack}
We address reward hacking at two levels: the tasks admitted to the training pool and
the incentives created by the reward function.

\paragraph{Data-side filtering: rejecting exploitable tasks.}
Our primary defence is to reject tasks that can be gamed before they enter the training
pool. The \strategypool pool is selected through a semantic audit with DeepSeek-V4-Pro.
Tasks receive a \code{hard\_reject} for any of four problems: hidden requirements, test
leakage, solution shortcuts, or verifiers too weak to validate the task's goal
(Section~\ref{sec:data}). This filtering aims to remove opportunities to earn reward
without accomplishing the intended task, and is our only defence directed at the
exploitability of the tasks themselves.

\paragraph{Reward-side design: monitoring turn growth and fixing the scale.}
The test-count reward can create an incentive to prolong trajectories simply to pass
more tests, leading to uncontrolled growth in turn counts. We observed this behaviour
in 27B experiments and investigated various length-shaping variants~\cite{rst}. For the 122B run, we closely monitored turn counts and
sequence lengths and did not observe the same runaway growth
(Figure~\ref{fig:dynamics}). The default 122B run uses the plain test-count reward,
with length shaping disabled.

We also keep $S=20$ fixed globally, using the value selected from the \strategypool
statistics in Section~\ref{sec:reward-def}. With normalization by a per-batch maximum,
a single extreme sample could change the denominator and hence the reward scale for
the entire batch. A fixed denominator makes each trajectory's reward independent of
the other samples' passing counts, removing this route for manipulating the batch's
reward scale.

\paragraph{Data interaction: shuffling is part of the reward design.}
\label{sec:reward-data}
The \strategypool pool is materialized in quality-rank order. Without shuffling a
sequential cursor would sweep from best to worst, at the batch size of 512 tasks would be
drawn from one narrow quality band and the reward distribution the critic sees would
drift monotonically over the epoch. 
\section{Results}
\label{sec:results}

\subsection{Benchmark and Harness}
\label{sec:benchmark}

We evaluate on three held-out suites, none of which contributes tasks to any training pool,
so that breadth, horizon length and raw difficulty are measured separately.

\begin{itemize}
  \item \textbf{\tbtwoone} \citep{terminalbench} is our primary held-out benchmark for
    terminal agent capability: 89 tasks graded by execution and reported as the fraction
    resolved. It supersedes \tbtwozero, whose instabilities hindered reproducible
    evaluation and underestimated benchmark performance.
  \item \textbf{\lhtb (LHTB)}
    \citep{li2026longhorizonterminalbenchtestinglimitsagents} is a suite of 46 hard,
    reproducible tasks across nine categories, designed to resist memorization,
    shortcutting and reward hacking. Every task pays continuous partial credit instead of
    binary pass or fail, so we report average reward rather than a resolved rate.
  \item \textbf{Terminal-Bench Hard (TBH)} \citep{rst} is a 100-task evaluation set 
    and reported as the fraction resolved. It probes an
    independently constructed and harder task distribution, and is distinct from the Hard
    difficulty group inside \tbtwoone.
\end{itemize}

All three suites run under one configuration. The agent is \terminus, hosted by \harbor: a
structured tool-call loop in which each assistant turn issues shell commands and each tool
turn returns terminal output, permitting at most 60 turns under an agent wall of 3600\,s and
a verifier wall of 900\,s. Explicit thinking is disabled in the chat template, proactive
context compaction is enabled so the agent summarizes its own history once the context
reaches a configured threshold, and every trial receives a freshly created cloud sandbox of
10\,GiB disk that is destroyed on completion. Decoding uses a context window of 96{,}000
tokens at temperature 0.1, top-$p$ 0.95 and top-$k$ 20, with 3 attempts per task. The
evaluation-total timeout is set to 18{,}000\,s. Two
starting checkpoints appear across our campaigns, the {base} model at 43.8\% and an SFT
checkpoint denoted {RST} at 49.4\%, the latter produced by rejection-sampling-style
fine-tuning outside our scope.

Training and evaluation share one software stack. We build our distributed training and
rollout framework on top of slime v0.3.0, with \megatron-Core v0.16.0rc0 and
Transformer Engine v2.10.0 on the training side, utilizing \sglang v0.5.12.post1
together with sglang-kernel v0.4.2.post2 and DeepEP v1.2.1 as the rollout inference
engine with vendor-specific synchronization and KV-cache optimization patches. Trials are
executed through \harbor v0.7.0 against \daytona sandboxes version v0.168.0. The
underlying runtime is PyTorch 2.11.0 with CUDA 12.9 and NCCL 2.28.9, and orchestration uses
Ray 2.55.1.

\subsection{Main results}

\begin{wrapfigure}{r}{0.3\linewidth}
  \centering
  \vspace{-2\intextsep}
  \includegraphics[width=\linewidth]{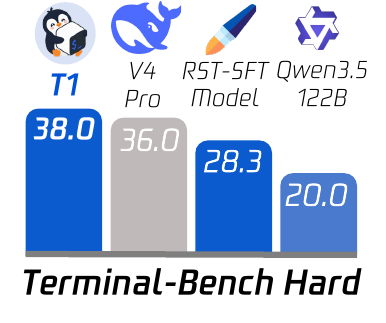}
  \caption{Terminal-Bench Hard resolved rate: \Tone outperforms the RST-SFT checkpoint by 9.7 points and the base model by 18.0 points.}
  \label{fig:tb_hard_results}
  \vspace{-2\intextsep}
\end{wrapfigure}

As shown in Table ~\ref{tab:frontier}, we evaluate the model on Terminal-Bench 2.1 and Long-Horizon Terminal-Bench (LHTB). Shown in Figure ~\ref{fig:tb_hard_results}, we evaluate the model on Terminal-Bench Hard for more challenge terminal tasks.

\paragraph{Binary reward on \tmax, from the base model.}
Our training is based on both critic model and actor model as Qwen3.5-122B-A10B without warm-up, using Tmax-15k as our training dataset.
This campaign improves the base model early, from 43.8\% to 47.2\% at iteration 30, but
never reaches the SFT checkpoint's 49.4\%. Its lasting contribution is the trained critic
of Section~\ref{sec:critic}.

\paragraph{Dense reward on the unfiltered \rstpool, from the SFT checkpoint.}
Our training is based on the warm-up critic model and the RST-SFT model as training initial actor model on RST-38k as our training dataset. The model reaches immediately above RST-SFT Model, reaching 59.9\% at iteration 70.


\paragraph{Dense reward on \strategypool, from the SFT checkpoint after Critic
Warm-Up.}
This is the production run of the present report: PPO at batch 512 with 84k context length,
routing replay enabled, and a critic obtained by Critic Warm-Up over \tmax for one epoch. 
Our training is based on the warm-up critic model and the RST-SFT model as training initial actor model on T1-15k as our training dataset.
At step 110 repeated evaluations reaches the best performance of \textbf{64.0\%}. 



\begin{table}[t]
\centering
\small
\caption{Performance Comparison. The upper block lists rows evaluated under the same
harness as ours, while the lower block gives selected public leaderboard rows obtained
under other harnesses and shown for context only, since harness choice materially changes
scores: Claude Opus 4.6 scores 70.1 under Claude Code against 63.8 under \terminus. Sizes
are as recorded in the evaluation workbook, and a solidus denotes a size not recorded.
The rightmost column reports average reward on Long-Horizon Terminal-Bench (LHTB).}
\label{tab:frontier}
\begin{tabular}{@{}llcc@{}}
\toprule
Model & Size (total-active) & \tbtwoone & LHTB \\
\midrule
\multicolumn{4}{@{}l}{\emph{Same harness (\harbor/\terminus):}} \\
Claude Opus 4.7 & / & 66.1 & -- \\
Claude Opus 4.6 & / & 63.8 & -- \\
Muse Spark & / & 62.2 & -- \\
Hy3-Preview & 295B-A21B & 58.0 & -- \\
DeepSeek V4 Flash (high) & 295B-A21B & 56.9 & -- \\
Kimi-K2.5 & 1040B-A32B & 56.4 & -- \\
Minimax M2.7 & 229B-A10B & 55.4 & -- \\
GPT-5.4 & / & 54.8 & 27.2 \\
Gemini 3 Flash & / & 54.2 & -- \\
Claude Sonnet 4.6 & / & 51.5 & 37.3 \\
\ourmodel (base) & 122B-A10B & 43.8 & 18.9 \\
\textbf{RST-SFT Model} & 122B-A10B & 49.4 & 23.6 \\
\textbf{\ourmodel + RL (Tmax-15k)} & 122B-A10B & 47.2 & 20.3 \\
\textbf{\ourmodel + RL (RST-38k)} & 122B-A10B &  59.9 & 25.4 \\
\rowcolor{metabg!8} \textbf{\Tone} & 122B-A10B & \bfseries 64.0 & \bfseries 27.9 \\
\midrule
\multicolumn{4}{@{}l}{\emph{Other harnesses (public leaderboards, context only):}} \\
GPT-5.3-Codex (Codex CLI) & / & 79.1 & 21.5 \\
GPT-5.4 (Codex CLI) & / & 77.3 & 27.2 \\
Claude Opus 4.6 (Claude Code) & / & 70.1 & -- \\
Gemini 3.1 Pro (Gemini Code) & / & 67.1 & 27.9 \\
GLM-5.1 (Claude Code) & 750B-A40B & 58.7 & 26.7 \\
\bottomrule
\end{tabular}
\end{table}

\begin{figure}[t]
  \centering
  \includegraphics[width=\linewidth]{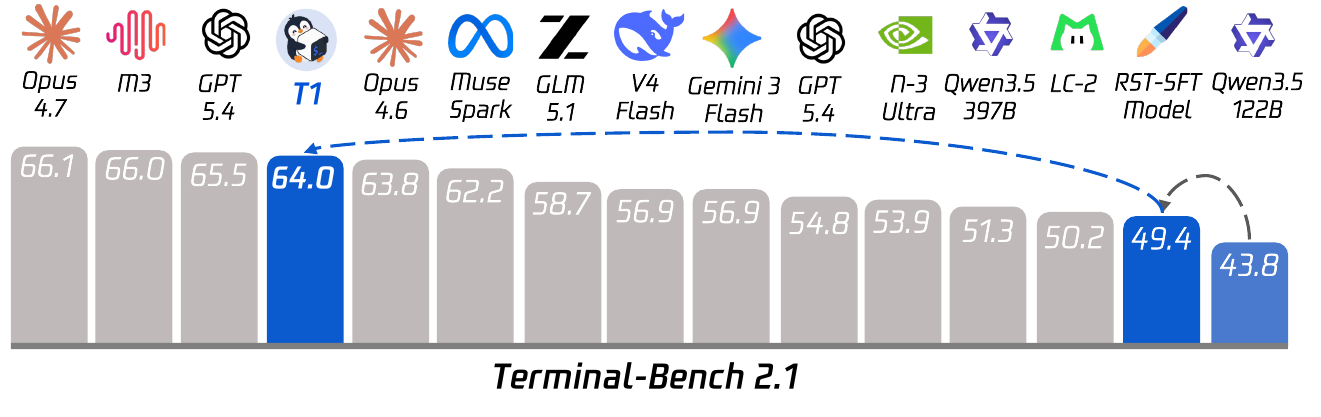}
  \caption{\tbtwoone standing of \Tone against contemporary frontier and open-weight
  models. Blue bars trace our own pipeline: the \ourmodel base model (43.8), the
  \rstpool SFT checkpoint we initialize from (49.4), and \Tone after RL (64.0); the
  dashed arrows mark the two stages, a gain of 5.6 percentage points from SFT and a further
  14.6 percentage points from reinforcement learning. Grey bars are the comparison models. With
  10B active parameters \Tone ranks fourth overall and ahead of Claude Opus 4.6, and is
  the only model in the leading group that reaches that band from a sub-50 starting
  point.}
  \label{fig:performance-raise}
\end{figure}

Under the same Terminus-2 harness and Daytona sandbox backend, reported in Table~\ref{tab:frontier}, the RL
checkpoint sits above GLM-5.1, Hy3-Preview, DeepSeek-V4-Flash, Kimi-K2.5, Minimax M2.7,
GPT-5.4 and Claude Sonnet 4.6, and within two points of Claude Opus 4.6, while using 10B
active parameters. We attach the standard caveat that harness and constraint choices
materially move these numbers, as discussed below, and that our model was RL-trained for
exactly this harness whereas the frontier models were not.

Figure~\ref{fig:performance-raise} puts that standing next to the trajectory that
produced it, which is the part a leaderboard row hides. Our three blue bars are the same
model at three stages: \ourmodel base at 43.8, the \rstpool SFT checkpoint at 49.4, and
\Tone at 64.0. Read left to right, the RL stage moves the model past nine of the
comparison systems in a single step. Supervised fine-tuning alone leaves it second from
last, below every comparison entry in the chart, while the same weights after RL
post-training land fourth overall and above Claude Opus 4.6. The contrast in step sizes is
the substantive claim: SFT contributes 5.6 percentage points and RL a further
14.6 percentage points, so
roughly three-quarters of the total distance from base to final is earned by
reinforcement learning on terminal tasks rather than by imitation of demonstrations. It is
also worth noting what the bars do not encode. The models above and immediately below us
are dense or far larger sparse systems, whereas \Tone reaches this band with 10B active
parameters, which restates the efficiency argument of Section~\ref{sec:intro} in
benchmark terms.

\begin{figure}[t]
  \centering
  \includegraphics[width=0.73\linewidth]{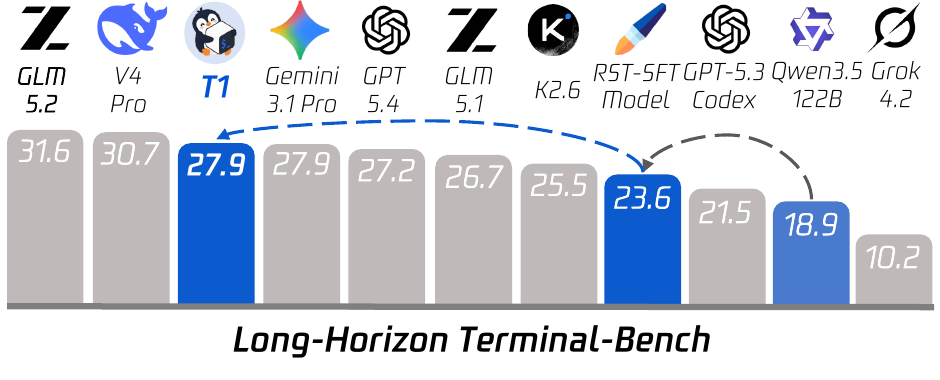}
  \caption{Performance on \lhtb. The base model, RST-SFT checkpoint, and
  \Tone score 18.9, 23.6, and 27.9, respectively. RL adds 4.3 average-reward
  points over SFT, and \Tone matches Gemini-3.1-Pro among the models shown.}
  \label{fig:lhtb}
\end{figure}
\paragraph{Transfer to longer horizons.}
\lhtb tests whether the gains on \tbtwoone extend to tasks that require more
sustained interaction with the environment. Figure~\ref{fig:lhtb} shows a consistent
improvement across our three checkpoints: the base model scores 18.9, SFT raises this
to 23.6, and RL reaches \textbf{27.9}. The RL stage therefore adds
\textbf{4.3 average-reward points} over SFT, an 18.2\% relative improvement; the
full pipeline gains 9.0 points over the base model. The improvement continues beyond
the earlier \strategypool checkpoint at iteration 70 (25.5) and the
\rstpool RL model (25.4) in Table~\ref{tab:frontier}, indicating that the stronger
\tbtwoone result is accompanied by progress on the longer-horizon evaluation.

Against the comparison models in Figure~\ref{fig:lhtb}, \Tone matches
Gemini-3.1-Pro at 27.9 and exceeds GPT-5.4 (27.2), GLM-5.1 (26.7), and Kimi-K2.6
(25.5), while GLM-5.2 (31.6) and DeepSeek-V4-Pro (30.7) remain ahead. These
results place a model with 10B active parameters in a competitive band on this
evaluation. The comparison is not uniform across benchmarks: Claude Sonnet~4.6,
which scores below \Tone on \tbtwoone, records 37.3 on LHTB in
Table~\ref{tab:frontier}. Longer-horizon performance therefore warrants a separate
evaluation rather than being inferred from the \tbtwoone ordering alone. We report
LHTB in its average-reward units, separately from the resolved percentages of the
other benchmarks.



\paragraph{Generalization to harder terminal tasks.}
Figure~\ref{fig:tb_hard_results} reports the complementary evaluation on
Terminal-Bench Hard (TBH). \Tone resolves \textbf{38.0\%} of tasks, compared with
28.3\% for the SFT checkpoint and 20.0\% for the base model. RL contributes
\textbf{9.7 percentage points} beyond SFT, a 34.3\% relative improvement, while the full
pipeline gains 18.0 percentage points over base. The final score also exceeds
DeepSeek-V4-Pro at 36.0\% by 2.0 percentage points. Together with the LHTB results, this
supports the view that
terminal-agent post-training improves performance as both task difficulty and
interaction horizon increase. The comparison concerns the complete training recipe;
it does not isolate the contribution of dense reward from the data pool or the
stabilization mechanisms.


\subsection{Training dynamics}
\label{sec:dynamics}

\begin{figure}[t]
  \centering
  \includegraphics[width=0.95\linewidth]{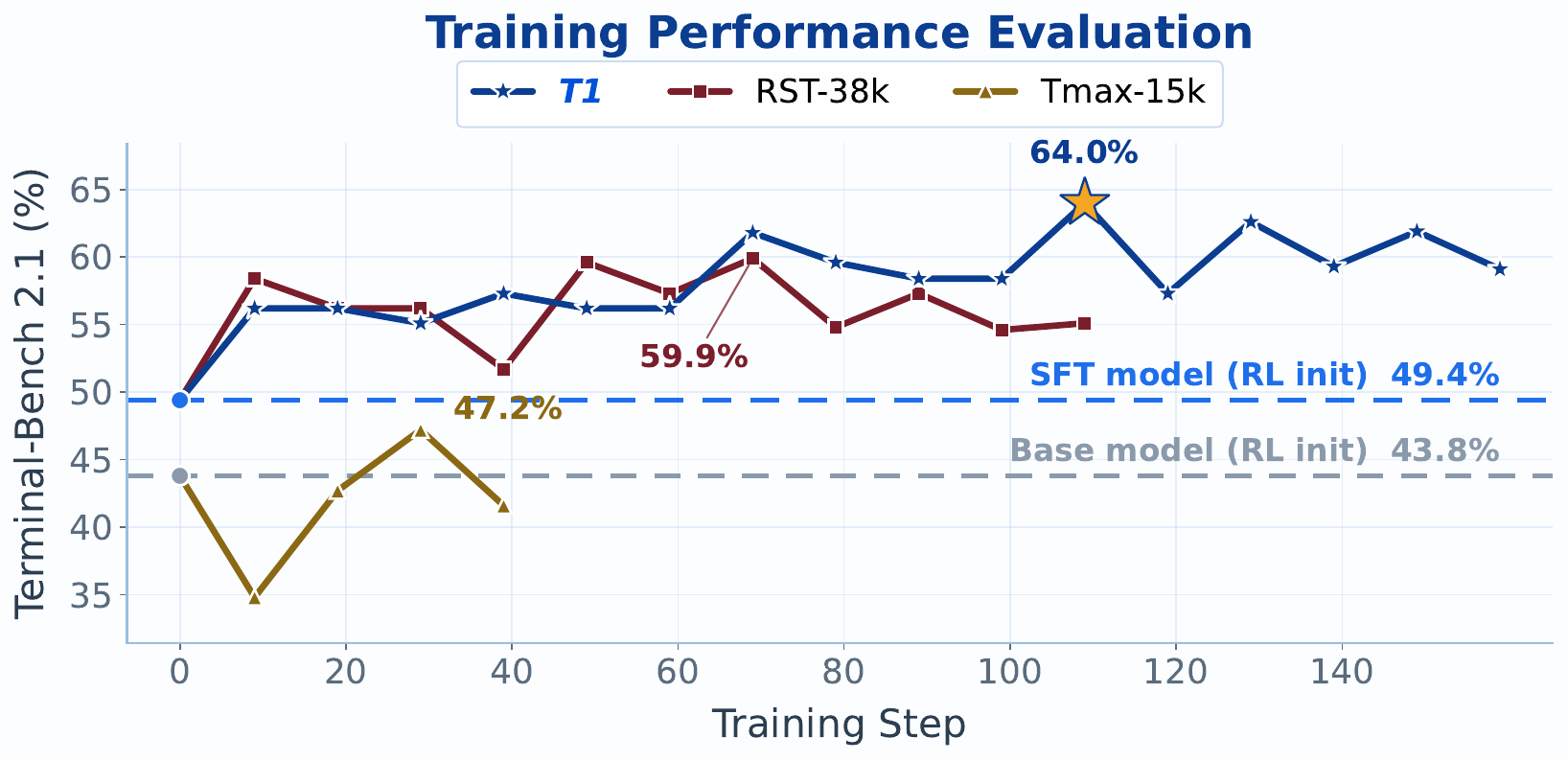}
  \caption{Held-out \tbtwoone resolved rate over the production dense-reward run.
  Filled markers are the evaluated checkpoints (every 10 rollout steps), each annotated
  with its score; the starred point is the peak. The dashed rules are
  the two fixed anchors, namely the base model at 43.8\% and the SFT checkpoint the run
  is initialized from at 49.4\%. The very first evaluated checkpoint already clears SFT
  by 6.8 percentage points, no evaluated checkpoint ever falls back to the initialization,
  and the peak at step 110 reaches 64.0\%, which stands 14.6 percentage points above SFT
  and 20.2 percentage points above base.}
  \label{fig:eval-dynamics}
\end{figure}

Figure~\ref{fig:eval-dynamics} traces the held-out \tbtwoone resolved rate of \Tone as a
function of training step, evaluated every ten rollout steps against the two fixed anchors
of the run, namely the base model at \textbf{43.8\%} and the SFT checkpoint at
\textbf{49.4\%} from which the run is initialized. The first evaluated checkpoint already
reaches \textbf{56.2\% at step 10}, exceeding that initialization by \textbf{6.8 percentage
points}, which is consistent with the reward curve of Figure~\ref{fig:reward-curve} rising
fastest over the same interval. Steps 20--60 then hold a band of \textbf{55.1--57.3\%}, a
spread of roughly three tasks out of 89 that is comparable to the variation observed across
repeated evaluations of a single checkpoint, and no evaluated checkpoint returns to the
initialization. Two further increases arrive late in the run, \textbf{61.8\% at step 70} and
\textbf{64.0\% at step 110}, the latter standing \textbf{14.6 percentage points} above SFT
and \textbf{20.2 percentage points} above the base model. Both coincide with the region
where explained variance in Figure~\ref{fig:critic-ev} is highest, which indicates that the
largest usable actor improvements occur only after the critic is well calibrated.

\begin{figure}[t]
  \centering
  \includegraphics[width=\linewidth]{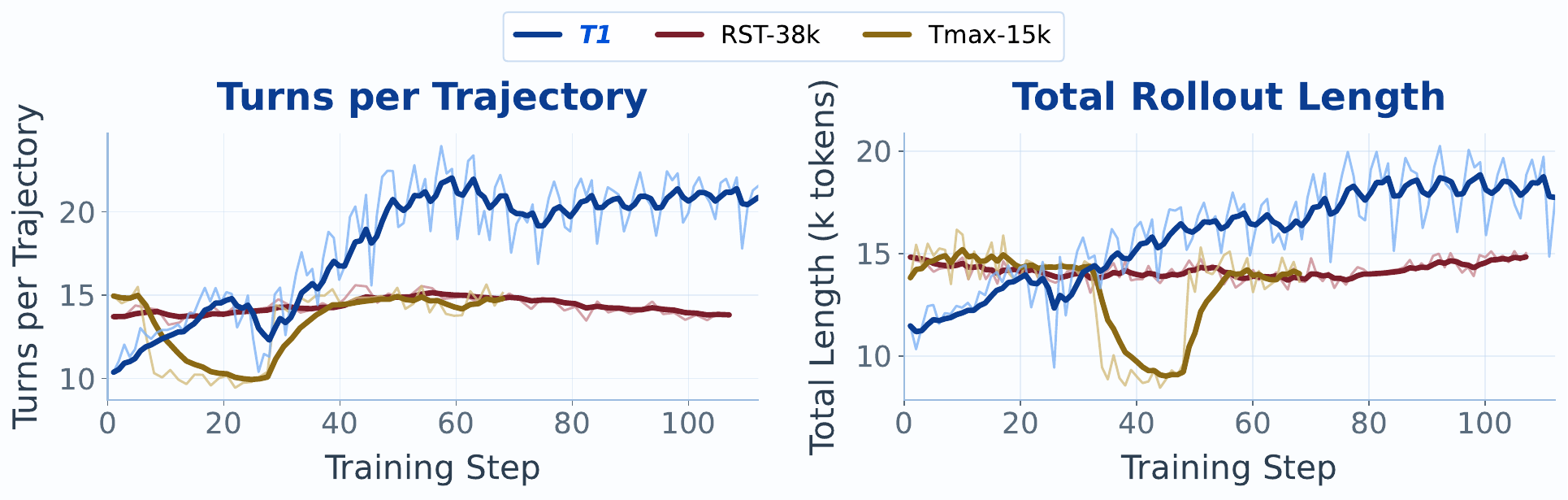}
  \caption{Behavioral dynamics of the dense-reward run. \figleft{} average tool-call
  turns per trajectory. \figright{} total sequence length per trajectory. Faint lines are
  per-step values, bold lines an EMA ($\alpha{=}0.25$). Both roughly double over
  training (turns $10.4 \to 20.9$, length $11.5\text{k} \to 17.8\text{k}$ tokens) and
  then flatten, rather than growing without bound.}
  \label{fig:dynamics}
\end{figure}

Figure~\ref{fig:dynamics} reports the behaviour that produces those scores. Average
tool-call turns per trajectory approximately double, from \textbf{10.4} to a plateau near
\textbf{20.7}, and total sequence length follows from \textbf{11.5k} to roughly
\textbf{18.3k} tokens. Two properties distinguish this trend from length hacking. The growth
is bounded, since both curves flatten after step 60 and the final third of training
contributes almost nothing, whereas the additive length penalty produced unbounded growth
beyond \textbf{50 turns}. The plateau is concurrent with the reward and benchmark gains, which indicates that
additional turns yield additional passing assertions rather than padding. Turn growth
nonetheless contributes to the timeout failures examined in Section~\ref{sec:case-study}.

\subsection{Case study: detailed evaluation across benchmarks}
\label{sec:case-study}

\begin{wrapfigure}{r}{0.49\linewidth}
  \centering
  \vspace{-1\intextsep}
  \includegraphics[width=\linewidth]{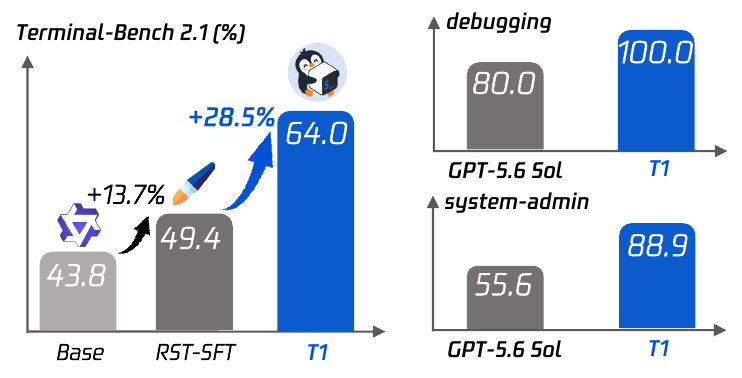}
  \caption{\figleft{} \tbtwoone progression across the three checkpoints of our
  pipeline, from base through SFT to \Tone. \figright{} per-domain comparison against
  GPT-5.6 Sol on the two subsets that depend on multi-step reasoning and tool
  use.}
  \label{fig:analysis}
  \vspace{-2\intextsep}
\end{wrapfigure}

\paragraph{Where the aggregate gain comes from.}
Figure~\ref{fig:analysis} decomposes the improvement from the base model to
\Tone. SFT raises the resolved rate from 43.8\% to 49.4\%, a gain of
5.6 percentage points, and RL adds a further \textbf{14.6 percentage points} to reach
64.0\%. Measured against each stage's initialization, these are 12.8\% and 29.6\% relative
improvements, respectively. RL thus accounts for 72.3\% of the total gain of
20.2 percentage points from base
to final, making it the larger contributor in this training pipeline.

The right panel highlights two domains where \Tone is particularly strong.
On debugging, it reaches \textbf{100.0\%} against 80.0\% for GPT-5.6 Sol;
on system administration, it reaches \textbf{88.9\%} against 55.6\%, gains of
20.0 and 33.3 percentage points, respectively. Both domains require the agent to inspect
environment state, act on a working hypothesis, and revise its approach from
execution feedback. Evaluate through \citet{wang2026harnesshandbookmakingevolving}, their results are consistent with the capabilities exercised
by our terminal-agent RL loop. The subsets are small, however: debugging contains
five tasks and system administration nine, so the domain scores describe specific
strengths within this evaluation rather than establishing an overall ordering
between the two models.



\begin{figure}[t]
  \centering
`  \includegraphics[width=\linewidth]{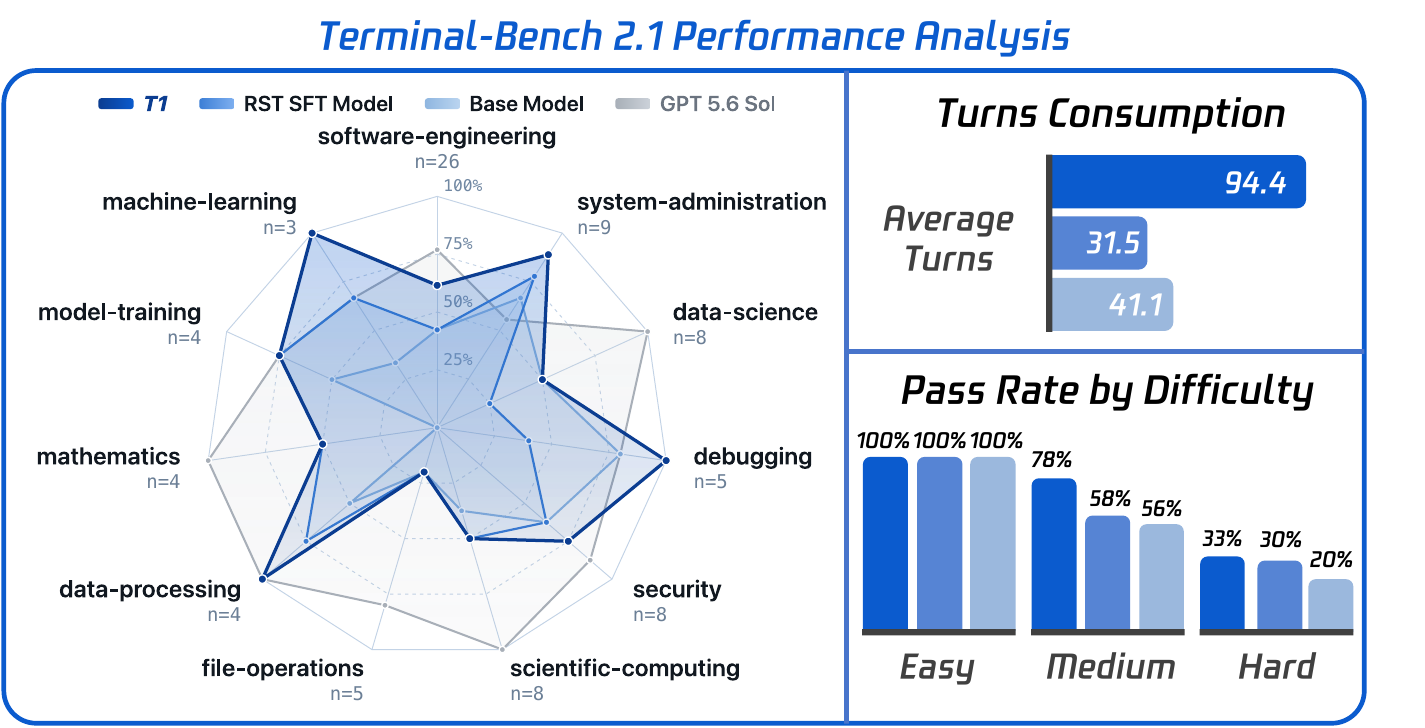}
  \caption{
  \tbtwoone performance by domain, interaction length, and difficulty.
  \figleft{} resolved rates across 11 domains, with task counts on each axis;
  GPT-5.6 Sol is included as an additional comparison. \figtop{} average turns
  for \Tone, RST-SFT, and base: 94.4, 31.5, and 41.1.
  \figbottom{} resolved rates for the same three checkpoints: all score 100\%
  on Easy tasks, while Medium scores are 78/58/56\% and Hard scores are
  33/30/20\%. Difficulty percentages are rounded as displayed in the figure.}
  \label{fig:category_difficult}
\end{figure}


\paragraph{Domain gains are uneven.}
The full breakdown in Figure~\ref{fig:category_difficult} shows that the advantage
extends beyond the two highlighted domains. \Tone also resolves all tasks in
data processing and machine learning, although these categories contain only four
and three tasks, respectively. The gains are not universal: GPT-5.6 Sol remains
ahead on data science, scientific computing, and mathematics, while file operations
remain difficult for our model. This pattern points to domain-specific headroom
that the aggregate resolved rate obscures. The concentration of \strategypool
in command-line engineering work (Section~\ref{sec:data}) provides one plausible
explanation for these differences, but the category comparison alone does not
separate data coverage from reasoning and tool-use limitations.

\paragraph{Most of the difficulty-level gain is on Medium tasks.}
All three checkpoints reach 100\% on the Easy group, leaving no measured headroom
there. On Medium tasks, \Tone reaches \textbf{78\%}, compared with 58\% for
SFT and 56\% for base: gains of approximately 20 and 22 percentage points. On Hard tasks,
the corresponding scores are \textbf{33\%}, 30\%, and 20\%, so the RL-stage
gain is smaller at approximately 3 percentage points. These rounded group scores show that
RL's improvement over SFT is largest on Medium tasks, while the hardest group
remains a substantial source of failures. The 33\% here and the 38.0\% on TBH
refer to different evaluations and should not be compared as successive checkpoints.

\paragraph{Higher success comes with more interaction.}
In the evaluation summarized by Figure~\ref{fig:category_difficult}, \Tone uses
94.4 turns on average, compared with 31.5 for SFT and 41.1 for base, approximately
$3.0\times$ and $2.3\times$ as many turns. This is consistent with a policy that
sustains longer attempts, but the aggregate statistics do not establish whether
the additional turns are productive on individual tasks. In particular, a large
increase in turns accompanies only a modest improvement on the Hard group.
These are evaluation trajectory lengths, distinct from the training averages in
Section~\ref{sec:dynamics}; they measure a cost of the resulting policy as well
as its capacity for extended interaction.

\paragraph{Failure cases expose inefficient search.}
The project case study against GPT-5.6 Sol makes this cost concrete. On six
representative unsolved tasks, \Tone spends between 164 and 473 turns, compared
with 6--40 for GPT-5.6 Sol, and times out; three further tasks fail on sandbox
errors. Only four failures reach the 500-turn evaluation ceiling, while eleven
record partial sub-test passes. The comparison also uses different resource
settings: our maximum input is 56k tokens against 120k, maximum output is
8{,}192 against 32{,}768, and explicit thinking is disabled against medium reasoning
effort. These differences limit attribution of the failures to model capability
alone. Together, the cases suggest that longer interaction is useful only when
paired with effective diagnosis, recovery, and stopping decisions; increasing
the turn budget alone would not address the observed failure modes.
Appendix~\ref{app:case-study} examines the complementary direction, contrasting
two tasks that \Tone resolves and its SFT initialization does not, to show which
behavioural changes the RL stage is responsible for.

\section{Infrastructure}
\label{sec:infra}

To maximize throughput, we decouple training and inference across disjoint accelerator pools in our pipeline using asynchronous reinforcement learning based on slime as our basic training framework \citep{slime}. While this overlap substantially improves training efficiency, it introduces severe infrastructure bottlenecks. Three challenges dominate: co-locating two 122B actor-critic parameter sets under strict device limits, managing hour-scale steps, and scaling out concurrent external sandboxes. We resolve each via principled resource and liveness models rather than heuristic tuning. Throughout, $G$ is the device count, $M$ the per-device memory budget, and a \emph{sharding plan} $\Pi$ assigns degrees $(\tau, \rho, \kappa, \delta)$ subject to $\tau\rho\kappa\delta = G$.


\subsection{A capacity model for co-resident actor--critic pairs}
\label{sec:infra-memory}

Because the two networks time-multiplex one device set through forced offload
(Section~\ref{sec:critic}), each must independently satisfy $M$, so memory rather than
arithmetic throughput is the binding constraint. The first joint update on our initial plan
aborted inside the critic's recurrent forward, requesting a further allocation of well under
one percent of $M$ on a device already resident above $95\%$ of its budget, after
critic-only steps had run cleanly. Rather than
search the plan space, we calibrated a closed-form per-device footprint, with
$P_{\mathrm{e}} = 116.0$B expert and $P_{\mathrm{d}} = 5.86$B dense parameters:
\begin{equation}
  \label{eq:capacity}
  b(\Pi, T)
  = \underbrace{\tfrac{6}{\rho}\!\left(\tfrac{P_{\mathrm{e}}}{\eta} + P_{\mathrm{d}}\right)}_{\text{weights} + \text{gradients}}
  + \underbrace{\tfrac{12\,P_{\mathrm{e}}}{G} + \tfrac{12\,P_{\mathrm{d}}}{\rho\,\delta}}_{\text{optimizer state}}
  + \underbrace{\alpha\,\tfrac{T}{\kappa} + \tfrac{4\,|\mathcal{V}|\,T}{\kappa}\,\mathbf{1}[\text{terminal stage}]}_{\text{activations} + \text{output logits}}
  + b_0
  \;\le\; M ,
\end{equation}
with $\eta$ the expert-sharding degree, $T$ the token wall, $|\mathcal{V}| = 248{,}320$, a
measured per-token activation coefficient $\alpha$, and a small constant allocator residue
$b_0$ amounting to a few percent of $M$. Equation~\ref{eq:capacity} reproduced the observed
abort to within a fraction of a percent of $M$,
and its structure rather than its numeric value determines the plan.

\begin{proposition}[Expert optimizer state is plan-invariant]
\label{prop:plan-invariance}
The optimizer footprint of the expert parameters depends only on $G$, never on how
$\Pi$ distributes them.
\end{proposition}
\noindent Per-device expert parameters shrink as $1/\eta$, but the distributed-optimizer
shard count grows as $\tau\kappa\delta/\eta$; their product is exactly $G$, so the two
effects cancel. The consequence was the most expensive lesson of bring-up:
\emph{reducing} expert sharding to relieve memory pressure doubles resident weights and
gradients while leaving optimizer state untouched, so $\eta$ is pinned maximal. Tensor
parallelism beyond $\tau = 2$ is likewise inert, since the recurrent blocks are not
tensor-sharded and the sequence is re-gathered to full length before them, so both their
parameters and their activation peak replicate rather than divide. Pipeline and context
degrees are the only real levers, and the binding term at long horizons is the
fixed-point output projection on the terminal stage: at an $84$k wall, $\kappa = 2$ leaves a
logit buffer occupying roughly a third of $M$ and therefore exceeds it, whereas $\kappa = 4$
halves that share and admits the plan.

\begin{keyinsight}[title=\textbf{Key Insight: capacity scales out, not in}]
{\Tonebody Proposition~\ref{prop:plan-invariance} says the dominant memory term of a
sparse actor--critic pair is invariant to every redistribution of experts and can be
reduced \emph{only} by enlarging $G$. Feasibility is therefore a property of the device
budget rather than of a tuning search, and this is what makes the recipe portable across
device budgets spanning more than a factor of two: a plan is admitted analytically, before
the $15$ to $20$\,min initialization, instead of empirically after an hour-long step dies.}
\end{keyinsight}

\subsection{Context parallelism for the recurrent operator}
\label{sec:infra-cp}

Equation~\ref{eq:capacity} divides activations by $\kappa$, but the generic
context-parallel path is unsound for a recurrent scan, which had historically forced
$\kappa = 1$ and made the single longest trajectory the memory wall. We therefore
sequence-shard the recurrent operator natively: each rank evaluates its own token
segment and hands the carried state to its successor, so the scan remains sequentially
exact while its activation residency divides. One subtlety is load-bearing: the trainer
distributes context shards in an interleaved layout for gradient balance, whereas state
passing requires contiguity, so shards are re-laid out and the packed-sequence descriptor
records how many interleaved segments a sample contributes, a distinction that is
unrecoverable from cumulative offsets alone, since real and padding segments can present
identical divisibility.

Correctness is asserted rather than assumed: against a single-rank reference the forward
pass is bit-exact and gradients agree to $1.2\times10^{-7}$ at $T = 33{,}792$ and
$65{,}536$. Relative to $\kappa = 1$, recurrent activation peaks fall to roughly one half,
one quarter and one eighth at $\kappa \in \{2,4,8\}$, a near-linear reduction and precisely
what makes $84$k and $128$k walls admissible on an unchanged device budget. Two costs are
accepted knowingly. Since
$\tau\rho\kappa\delta = G$ is fixed, raising $\kappa$ consumes data parallelism: a batch
of $512$ at $\delta = 2$ becomes $256$ serially accumulated micro-batches per rank.
Second, the kernel's long-sequence algorithm selection must be pinned, since the library
silently switches implementations under inference-mode heuristics, decoupling trainer
numerics from the sampler and reintroducing the very engine gap that
Section~\ref{sec:stability} exists to close.

\subsection{Liveness under hour-scale steps}
\label{sec:infra-ft}

At our production device budget, where a single step takes approximately one hour, events
with per-hour probability $10^{-2}$ are per-run certainties, and the failures we observed
were not crashes but
\emph{indefinite waits}: a single unresponsive participant stalls a publication barrier that
spans the entire training partition, and the job dies of a downstream timeout whose message
names the barrier rather than the cause. Two principles proved sufficient. First, every
control-plane operation carries a bounded deadline, while bulk transfers deliberately do
not, since a multi-minute collective copy is legitimate whereas an unbounded teardown
request is a liveness hole. Second, detection must distinguish \emph{death} from
\emph{slowness}: deadlines inherited from single-node defaults misclassify benign
stragglers as failures, since a first-step kernel compilation or a cold-cache rank is
slow by construction. Probes therefore carry a generous initial grace of $600$\,s before
steady-state checks at a $30$\,s period and a $120$\,s deadline, and re-created collective
groups take deadlines calibrated to the regime of $30$ to $120$\,min rather than to
interactive latencies. Unresponsive replicas are reclaimed at the next publication
barrier, amortizing recovery into a synchronization point that already exists.
Checkpointing is deliberately asymmetric, saving weights only and every ten steps, which
trades resumability for cost on the understanding that a crash means restart rather than
resume.

\subsection{Environment concurrency and the straggler tail}
\label{sec:infra-sandbox}

Reward evaluation is an external service, so the rollout layer is an admission-control
problem. Environment instances are hosted by per-node auxiliary workers pinned to the
inference partition with a per-node concurrency cap ($35$ at batch $256$, $70$--$94$ at
batch $512$), keeping trial traffic off the coordinator, and creation is paced by a token
bucket at $6$\,s$^{-1}$ against a provider quota of $600$\,min$^{-1}$. Instances are
provisioned minimally, with a single core and a small memory and disk allotment, because
concurrency, not
per-instance capability, sets the achievable batch; reclamation must be explicit, since
deferred deletion that triggers only after suspension leaks instances until quota
exhaustion. Per-trial deadlines are layered ($5400$\,s remote, $3600$\,s agent, $900$\,s
verifier).

The scheduler oversamples: it launches $560$ trials for a batch of $512$, admits the
first $512$ to complete, and cancels the remainder. This bounds step time against a
heavy-tailed completion-time distribution, but the residual cost is structural. Once a
handful of trials remain, no admission decision can be made until one terminates, and
near-deadline trajectories are exactly those grinding through context compaction at
degraded decode rates; one episode spent $17$ minutes at $596$ trajectories collected
against $512$ groups required. The cancelled tail is not a uniform sample but
concentrates the hardest task families, so oversampling trades wall-clock determinism
against a selection bias quantified in Section~\ref{sec:limitations}.

\subsection{The publication barrier and rollout/training balance}
\label{sec:infra-sync}

Updated parameters are published to every inference replica once per step over dedicated
collectives, with generation quiesced first so that no request straddles two versions;
this is the barrier that makes the staleness of Appendix~\ref{app:async} exactly one
update rather than an unmodeled random variable.

Because the two partitions are disjoint and pipelined one step deep, step time is $\max$
rather than sum, which turns device allocation into a genuine optimization. Let $n$
devices serve inference and $G - n$ train; with per-replica throughput approximately
additive,
\begin{equation}
  \label{eq:balance}
  T_{\mathrm{step}}(n) \;=\; \max\!\Big(\underbrace{\tfrac{R}{n}}_{\text{rollout}},\;
  \underbrace{\tfrac{C}{G-n}}_{\text{training}}\Big),
  \qquad
  n^{\star} = \arg\min_n T_{\mathrm{step}}(n) \;\text{ attained at equality.}
\end{equation}
Measurement confirms the shape: in an earlier configuration, moving from one to three
inference replicas cut generation from $58.9$ to $19.6$\,min (single-replica rate $4.35$
groups/min), reducing a step of approximately $60$\,min to approximately $23$\,min, a factor
of $2.6$ with no
additional hardware, purely by rebalancing $n$. The production plan sits near the balance
point and spends about half of the device budget on inference, hiding approximately
$16$\,min of generation under approximately $41$\,min of training compute at $\delta = 2$,
for a step of approximately $57$\,min. That the terms are deliberately \emph{unequal} is the
caveat: generation slack
absorbs the straggler tail of Section~\ref{sec:infra-sandbox} without exposing it in
$T_{\mathrm{step}}$.

\begin{algobox}{One outer iteration. Generation for step $t{+}1$ overlaps the update at
step $t$, and the publication barrier is the only synchronization point between the two
partitions, which is what pins the behaviour policy to a single version.}{alg:infra-loop}
\Require plan $\Pi$; batch $B$; oversampling factor $1{+}\varsigma$
\State \textbf{assert} $b(\Pi, T) \le M$ and $\tau\rho\kappa\delta = G$
       \Comment{fail fast, before initialization}
\For{$t = 0, 1, \dots$}
  \State \textbf{inference partition:} admit $\lceil (1{+}\varsigma)B \rceil$ trials under
         the rate bound; collect the first $B$ to terminate; cancel the tail
  \State \textbf{training partition:} update the critic on $\mathcal{B}_t$, then the actor
         against its pre-update values (Section~\ref{sec:critic})
  \State quiesce generation; publish $\theta_{t+1}$; reclaim dead replicas; rotate the
         behaviour snapshot \Comment{single barrier}
\EndFor
\end{algobox}

\section{Related Work}
\label{sec:related_work}

\paragraph{Data synthesis and training for terminal agents.}
Expert-authored terminal tasks pair an instruction with a container and an executable
verifier \citep{terminalbench}, but manual authoring does not reach training scale, so
recent work synthesizes environments along three routes. \emph{Repository-derived}
methods recover workspaces from real development histories and reuse the accompanying
tests \citep{swegym,r2egym,terminaltraj}; \emph{perturbation} methods inject faults into
working repositories or CLI workspaces \citep{swesmith,cligym}; and \emph{task-conditioned
synthesis} generates the instruction, environment and verifier jointly from categories,
capability taxonomies or skill graphs \citep{endlessterminals,termigen,nemotronterminal,cliuniverse,tmaxdata,skillsynth}.
Because tasks synthesized from scratch saturate quickly against frontier models, a second
line makes the pool itself adaptive: RST recursively re-seeds validated task bundles to
lengthen horizons \citep{rst}, SETA and environment evolution raise difficulty generation
by generation \citep{seta,envevolution}, CalibForge calibrates against solver feedback
\citep{calibforge}, and Terminal-Universe reconstructs executable workspaces from recorded
trajectories before re-querying them across workspaces and dialogue rounds
\citep{terminaluniverse}. On the training side, most of these pipelines are validated by
supervised fine-tuning alone, and the reinforcement-learning evidence is confined to
comparatively small dense policies optimized against binary outcomes, with reported gains
over the SFT checkpoint often within a few points \citep{endlessterminals,otagent,tmaxdata,rome}.

\paragraph{Stable reinforcement learning under training--inference mismatch.}
Modern RL stacks sample with an inference engine and differentiate with a training engine
\citep{sglang,megatron,slime}, so differing kernels, numerics and parallelism make the
sampler's log-probabilities diverge from the trainer's even at zero staleness, corrupting
the importance ratios rather than merely dating them \citep{mis}. Sparse models amplify
this, since one update flips roughly a tenth of the activated experts for the same prefix
and token ratios then compare two different subnetworks \citep{gspo}. Existing remedies
reweight, mask, or reshape the objective. Truncated importance sampling corrects the ratio
in place \citep{tis}, though token-level correction leaves the induced state distribution
biased \citep{mis}; IcePop instead drops gradients outside a fixed two-sided band
\citep{icepop}, KPop replaces that band with a binary-KL acceptance region so exploratory
low-probability tokens are not over-masked \citep{kpop}, and SAT contracts only the
sign-selected clip endpoint on a self-calibrating staleness quantile \citep{sat}; GSPO and
DPPO act on the objective itself, moving the ratio to the sequence level or replacing ratio
clipping with a direct divergence estimate \citep{gspo,dppo}. Closest to us are methods
that remove the mismatch at its source: rollout routing replay reinstates the sampler's
expert selections in the backward pass \citep{r3}, and token-in-token-out makes the trainer
score exactly the identifiers the engine consumed and emitted \citep{tito}. 

\section{Lessons learned and what did not work}
\label{sec:lessons}

The failures that most changed our understanding of long-horizon RL concerned three
parts of the learning problem: what information the reward provides, what history the
agent can use, and which tasks reach the optimizer. Each can change the training
outcome while leaving the PPO update itself intact.

\subsection{Critic calibration does not guarantee policy improvement}
In the 27B pathfinding runs, increasing the critic learning rate brought the first
positive explained variance forward from step 50 to step 30, yet rollout reward
remained flat across the compared settings (Section~\ref{sec:critic}). The critic was
learning to predict returns more effectively without a corresponding improvement in
the reward earned by the actor. This redirected attention to the information supplied
by the tasks and their verifiers.

The critic learns to predict the return that the reward function assigns. When hard
tasks mostly receive the same zero outcome, improving that prediction cannot reveal
partial progress that the verifier never rewards. The task pool and reward granularity
determine which improvements are observable; critic calibration determines how well
the value baseline models the resulting returns. Our dense reward addresses the former
by making partial completion visible, while critic warm-starting and faster value
learning address the latter.

The practical lesson is to diagnose these requirements separately. Explained variance
is useful evidence about the value function, while passing tests, task completion, and
held-out evaluation establish whether the policy is improving. When calibration gets
better but reward does not move, task difficulty and verification granularity deserve
attention alongside further optimizer tuning. 

\subsection{Context management determines what the agent can learn}
An agent can become less efficient because it loses track of what it has already
done. In one campaign, a mismatch between the context budget and the requested summary
length caused 98.3\% of full-summary attempts to fail. The fallback retained only a
short fragment of the history. Agents repeated completed work, average turns rose from
22 to 30, and more trajectories reached their time limits. The run continued to produce
trajectories and rewards, but the agent was making decisions with degraded memory.

For a long-horizon agent, context compaction determines which earlier observations and
decisions remain available to the policy. Losing that information changes the effective
task presented at subsequent turns. Training can then spend capacity learning to cope
with avoidable information loss. Faithfully optimizing the collected trajectory does
not restore the history that was missing when its actions were chosen.

This makes turn growth alone an ambiguous diagnostic: it can reflect useful additional
work, reward-driven repetition, or a failure to preserve context. Before changing the
reward in response, we need to inspect what the agent remembers across compaction
boundaries and whether it repeats work whose results were lost. Applying a length
penalty to such trajectories would leave the memory failure unresolved.

\subsection{Rollout throughput changes the training distribution}
We deliberately mix task difficulties through shuffling, but the rollout collection
rule introduces another selection step. To bound batch latency, the scheduler launches
more trials than needed, accepts the first completed batch, and cancels the remaining
tail (Section~\ref{sec:infra-sandbox}). In one measured step, 48 of 561 submitted trials
were cancelled even though the accepted batch was full. A completed-batch counter
therefore concealed the loss of those training opportunities.

Completion time varies with task family, difficulty, and the agent's behaviour, so
selection by completion time can favour faster tasks. Shuffling the launch order does
not remove this bias: the distribution that reaches the optimizer also depends on
which trajectories survive collection. This is especially consequential for our dense
reward design. Partial progress on hard tasks is useful only if those trajectories
are retained long enough to be verified and used in an update. A richer reward cannot
recover supervision from a cancelled trajectory.

Throughput must therefore be assessed together with task retention, including which
families and trajectory lengths are being dropped. We retain oversampling for its
wall-clock benefit, while treating its effect on long-task coverage as an open
trade-off. Preserving unfinished trajectories through partial-rollout continuation is
one direction discussed in Section~\ref{sec:limitations}; its benefit to final task
performance remains to be established.


\subsection{Why PPO rather than a critic-free group baseline}
\begin{wrapfigure}{r}{0.45\linewidth}
  \centering
  \vspace{-0.6\intextsep}
  \includegraphics[width=\linewidth]{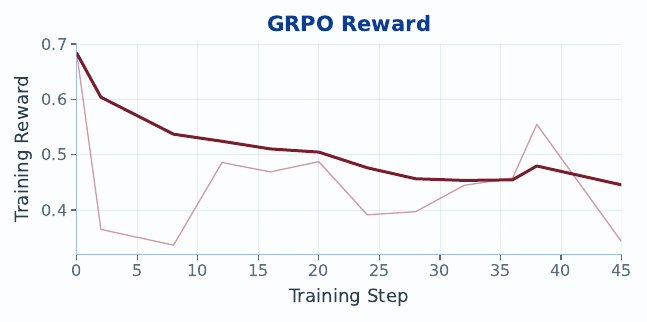}
  \caption{Rollout reward of the GRPO campaign on \strategypool under the same harness
  and reward as the production run. The faint line is the per-step mean and the bold
  line an exponential moving average at $\alpha{=}0.25$.}
  \label{fig:grpo-reward}
  \vspace{-0.4\intextsep}
\end{wrapfigure}

Before committing to PPO with a full-size critic, we trained GRPO on \strategypool under
the same harness, reward and device budget. The rollout reward in
Figure~\ref{fig:grpo-reward} fluctuates without a trend, and the held-out evaluation is
identical at the two checkpoints we scored, \textbf{51.7\%} at both step 10 and step 20,
which is the same 46 of 89 tasks resolved. Three properties of the group-relative
estimator account for this, and each is specific to long horizons.

\paragraph{The advantage degenerates precisely where signal is required.}
GRPO replaces the value baseline with the within-group statistics of $G$ trajectories drawn
from one task, $A_i = (r_i - \mu)/\sigma$, so a group whose members all fail, or all
succeed, contributes no gradient. On tasks hard enough to motivate the dense reward of
Section~\ref{sec:reward}, such groups are the common case, and the few that do vary
dominate the update. A single advantage is then shared by every token of a trajectory,
which removes the temporal credit assignment that GAE supplies.

\paragraph{Group size and task diversity compete for long-horizon tasks.}
Each task must be sampled $G$ times before it yields one advantage, so at fixed generation
capacity the number of distinct tasks per step falls by that factor, and the effective
batch narrows to a small set of repeatedly sampled tasks. PPO spends the same budget on
distinct tasks, because a learned value function provides the baseline that GRPO has to
purchase with repeated long-horizon rollouts.

\paragraph{The cost of those repeats is paid at the tail.}
A group closes only when its slowest member terminates, and terminal trajectories are
heavy-tailed in both turns and tokens, so the truncation that bounds step time for
independent trials cannot be applied inside a group without discarding the comparison it
exists to form. Generation therefore idles at the very configuration the estimator
requires. Taken together, the group baseline is a substitute for a value function whose
cost is modest in single-turn settings and grows with the horizon, whereas the critic's
cost is real but bounded, as Section~\ref{sec:critic} sets out.

\section{Limitations and future work}
\label{sec:limitations}

\paragraph{Conditioning is exact only up to re-tokenized history.}
\TITO guarantees exactness on the loss region, but in the re-tokenized case later turns
are trained conditioned on sampled identifiers where inference conditioned on the
harness's re-tokenized ones, which affects 2.6\% of tokens in the audited run.
Eliminating the residual would require the harness to carry token identifiers as the
source of truth for history, which we did not attempt.

\paragraph{No single-axis reward ablation yet.}
The evidence comparing binary against dense rewards is campaign-level, since the two runs
differ in reward, pool, initialization and routing replay at once. A binary run matched on
pool, initialization and routing replay would isolate the reward's contribution.

\paragraph{Verifier integrity is filtered, not enforced.}
The verifier executes in the agent-controlled sandbox with no runtime tamper
detection; defense currently rests on the audited task pool. In-sandbox integrity
checks (read-only test mounts, checksummed verifiers) are straightforward next steps.

\paragraph{The long tail is being paid for twice.}
Skim-oversampling drops the hardest trials from training, and the case study shows
remaining evaluation failures concentrate on exactly such tasks (hundreds of turns,
timeouts). Partial-rollout continuation, per-task budget respect , larger evaluation budgets, and
difficulty-aware scheduling are all in flight.

\paragraph{Data distribution.}
Failures concentrate in machine learning / data-science / scientific-computing categories that
\strategypool under-coversIn the future, we hope to explore more comprehensive data distributions to enable trained models to achieve versatility across various domain categories.

\paragraph{Staleness and asynchrony.}
The production loop is one-step asynchronous; the fully-asynchronous path exists in
the plugin but was not used for the 122B campaigns. Studies \citep{sat} on
Qwen3-30B-A3B (AIME24) show baseline GRPO/GSPO collapsing under staleness 8 and
identify combinations (GSPO+SAT+\Rthree, soft masking, TIS) that survive; porting the
winning combination to the terminal-agent setting is planned. The meeting record also
discusses critic \emph{value pretraining} on offline trajectories and
classification-based value losses (HL-Gauss) for step-0 explained variance; neither is
implemented in this codebase today.

\paragraph{Reporting gaps.}
Several bookkeeping items remain outstanding: the final optimizer-placement setting of
the production runs, the coverage of the length-adaptive advantage variant, the per-row
evaluation protocol labels, and the weight-synchronization timings. These are bookkeeping
matters rather than blockers.

\paragraph{Scope.}
Everything here is one model family, one agent harness and one benchmark. The
transferable claims are the mechanisms and their failure modes, namely token-faithful
trajectory construction, routing replay, critic scheduling and the operational
guardrails, rather than the specific numbers.

\bibliographystyle{plainnat}
\bibliography{references}

\clearpage
\beginappendix
\section{Notation}
\label{app:notation}

Tables~\ref{tab:notation-idx}--\ref{tab:notation-rl} collect every symbol used in the
paper. Sections~\ref{sec:mismatch}--\ref{sec:mismatch-effect} use exactly these
conventions. Three of them are worth stating explicitly up front, because they carry most
of the argument of Section~\ref{sec:stability}.

\begin{enumerate}
\item \textbf{Policy versions are indexed by the training step at which the weights were
  produced.} $\pi_t$ is the policy whose parameters are $\theta_t$, \ie the weights the
  trainer holds while performing update $t$; $\pi_{t-1}$ is the immediately preceding
  version. In the asynchronous loop the batch consumed at step $t$ was \emph{generated}
  under $\pi_{t-1}$, so the two symbols are never interchangeable
  (Appendix~\ref{app:async}).
\item \textbf{Superscripts $\mathrm{r}$ and $\mathrm{t}$ mark the \emph{engine}, not the
  version.} $\pi^{\mathrm{r}}$ denotes evaluation by \sglang (rollout) and
  $\pi^{\mathrm{t}}$ evaluation by \megatron (trainer). Engine and version are orthogonal
  axes: $\pi^{\mathrm{r}}_{t-1}$ and $\pi^{\mathrm{t}}_{t-1}$ are the \emph{same} weights
  under two implementations, which is precisely the discrepancy \TITO and \Rthree address.
\item \textbf{Bold lowercase is a token-indexed sequence; calligraphic uppercase is a set
  or an index collection.} Thus $\va_i$ is turn $i$'s completion and $\sI^{\ell}_j$ is the
  expert index set chosen at layer $\ell$ for position $j$.
\end{enumerate}

Two pairs of symbols are deliberately close and must not be conflated: the PPO clip
$\varepsilon$ versus the router-perturbation bound $\epsilon$, and the discount $\gamma$
versus the top-$k$ margin $\gamma^{\ell}_j$. Both distinctions are flagged in the tables.

\begin{table}[H]
\tablesize
\centering
\caption{Index conventions. These are fixed throughout the paper; in particular $t$ is
never a token index.}
\label{tab:notation-idx}
\begin{tabularx}{\linewidth}{@{}l l X@{}}
\toprule
\textbf{Symbol} & \textbf{Type} & \textbf{Meaning} \\
\midrule
$t$ & integer & \emph{Training step} / policy version. Never a token index. \\
$j$ & integer & \emph{Position} in the stitched token stream, $j \in [N]$. \\
$i$ & integer & \emph{Turn} within a trial, $i \in [T]$. \\
$\ell,\ e$ & integers & MoE layer $\ell \in [L]$ and expert $e \in [E]$. \\
$n$ & integer & Lookahead offset inside the GAE sum (Equation~\ref{eq:ppo_updates}). \\
\bottomrule
\end{tabularx}
\end{table}

\begin{table}[H]
\tablesize
\centering
\caption{Policy versions, engines, and optimization state (Appendix~\ref{app:async}).}
\label{tab:notation-ver}
\begin{tabularx}{\linewidth}{@{}l l X@{}}
\toprule
\textbf{Symbol} & \textbf{Type} & \textbf{Meaning} \\
\midrule
$\theta_t,\ \pi_t$ & params, policy
  & Actor parameters after update $t$, and the induced policy. $\pi_0$ is the SFT
    initialization. \\
$\pi_{t-1}$ & policy
  & The previous version; in the asynchronous loop, the version that \emph{generated} the
    batch consumed at step $t$. \\
$\gB_t$ & set
  & The batch consumed by update $t$; $\gB_t \sim \pi^{\mathrm{r}}_{t-\sigma}$
    (Equation~\ref{eq:async-loop}). \\
$\pi^{\mathrm{r}},\ \pi^{\mathrm{t}}$ & policy
  & Same weights evaluated by the sampler versus by the trainer.
    Superscripts compose with version subscripts: $\pi^{\mathrm{r}}_{t-1}$. \\
$\pi_{\mathrm{old}}$ & policy
  & Behaviour policy of the PPO ratio. Here it is $\pi^{\mathrm{t}}_{t-1}$, the previous
    version recomputed on the training side rather than read back from the sampler
    (Equation~\ref{eq:ppo-ratio-async}). \\
$\pi_{\mathrm{ref}},\ \theta_{\mathrm{ref}}$ & policy
  & Frozen reference for the KL diagnostic; routes with its own $\operatorname{TopK}_k$. \\
$\sigma$ & integer
  & Staleness, $\sigma = t - (\text{version that generated } \gB_t)$. Production runs use
    $\sigma = 1$. \\
$\phi_t,\ V_{\phi}$ & params, function
  & Critic parameters after update $t$, and the value function.
    $V_{\mathrm{old}} = V_{\phi_{t-1}}$ (Equation~\ref{eq:ppo_updates}). \\
$\eta_\theta,\ \eta_\phi$ & scalars
  & Actor and critic learning rates; $\eta_\phi = 30\,\eta^{\text{canon}}$,
    $\eta_\theta = 2\,\eta^{\text{canon}}$ with
    $\eta^{\text{canon}} = 5\times10^{-7}$ (Section~\ref{sec:critic}). \\
\bottomrule
\end{tabularx}
\end{table}

\begin{table}[H]
\tablesize
\centering
\caption{Trajectories, tokens, and MoE routing (Sections~\ref{sec:tito},
\ref{sec:r3}).}
\label{tab:notation-tok}
\begin{tabularx}{\linewidth}{@{}l l X@{}}
\toprule
\textbf{Symbol} & \textbf{Type} & \textbf{Meaning} \\
\midrule
$\gV$ & set & Vocabulary. \\
$\mathrm{enc},\ \mathrm{dec}$ & maps
  & Tokenizer and its inverse rendering. Token drift is exactly the failure of
    $\mathrm{enc} \circ \mathrm{dec} = \mathrm{id}$. \\
$T$ & integer & Number of turns in a trial. \\
$\gM_i$ & sequence & Message history presented to the template at turn $i$. \\
$\vp_i,\ \va_i$ & $\in \gV^{*}$
  & Turn $i$'s prompt token ids and sampled completion token ids. \\
$\tilde{\va}_i$ & $\in \gV^{*}$
  & The harness's re-encoding $\mathrm{enc}(\mathrm{dec}(\va_i))$; never enters the
    training stream (Equation~\ref{eq:retok-substitution}). \\
$\vq_i$ & $\in \R^{*}$ & Sampler log-probabilities for $\va_i$ (per turn). \\
$\vx,\ N$ & $\in \gV^{N}$, integer
  & The stitched training token stream and its length; $x_j$ is its $j$-th token. \\
$\vm,\ m_j$ & $\in \{0,1\}^{N}$
  & Loss mask; $m_j = 1$ iff $x_j$ was sampled by $\pi^{\mathrm{r}}_{t-1}$
    (Equation~\ref{eq:mask}). \\
$\vq,\ q_j$ & $\in \R^{N}$
  & The per-turn $\vq_i$ stitched onto the stream, $0$ where $m_j = 0$. \\
$\vu,\ \vv$ & $\in \gV^{*}$
  & Generic token sequences, used only to state the prefix relations. \\
$\Vert,\ \preceq,\ \ominus$ & operators
  & Concatenation; ``is a prefix of''; residual suffix after removing a prefix. \\
$\mathrm{drop}_s$ & operator
  & Removal of $s$ tokens from a sequence edge (Equation~\ref{eq:case-norm}). \\
$\vp_{i+1}[b\!:\!e]$ & slice
  & Half-open token span located by the generation-time offset map. \\
$\gA,\ \gD,\ \gP$ & sets
  & Partition of the $N$ stream positions into exactly-aligned, re-tokenized and
    placeholder positions (Equation~\ref{eq:audit}). \\
\addlinespace
$L,\ k,\ E$ & integers
  & MoE layers, top-$k$ width, experts per layer; $(48, 8, 256)$ for \ourmodel. \\
$d$ & integer & Model hidden size. \\
$\vh^{\ell}_j,\ \vy^{\ell}_j$ & $\in \R^{d}$
  & Layer-$\ell$ input and output at position $j$. \\
$\vs^{\ell}_j$ & $\in \R^{E}$
  & Router scores; $[\vs^{\ell}_j]_e$ is expert $e$'s score. \\
$f_e$ & map & Expert $e$'s feed-forward transform. \\
$\sI^{\ell}_j$ & $\subset [E]$
  & Selected expert set, $|\sI^{\ell}_j| = k$ (Equation~\ref{eq:moe}). \\
$\sI_j$ & tuple
  & $(\sI^{1}_j, \dots, \sI^{L}_j)$: the effective sub-network at position $j$. \\
$g^{\ell}_{j,e}$ & scalar
  & Gate weight applied to expert $e$; gradient flows through it. \\
$\gamma^{\ell}_j$ & scalar
  & Top-$k$ margin $[\vs^{\ell}_j]_{(k)} - [\vs^{\ell}_j]_{(k+1)}$. Distinct from the
    discount $\gamma$. \\
$\xi^{\ell}_j,\ \epsilon$ & vector, scalar
  & Cross-engine router-score perturbation and its bound
    $\|\xi^{\ell}_j\|_\infty \le \epsilon$. Distinct from the GAE residual $\delta_j$ and
    the PPO clip $\varepsilon$. \\
$\tR$ & $\in [E]^{\rho \times L \times k}$
  & Recorded routing tensor; $\tR[j,\ell,:]$ is the expert set used to predict position
    $j{+}1$ (Equation~\ref{eq:rows}). $\tR_i$ is turn $i$'s capture. \\
$\rho_i,\ \rho$ & integers
  & Routing rows per turn, $\rho_i = |\vp_i| + |\va_i| - 1$, and for the stitched stream,
    $\rho = N - 1$ (a turn's last token predicts nothing). \\
$\Pi$ & operator
  & Sharding composition applied identically to $\vx$ and $\tR$:
    $\Pi = \Pi_{\text{SP}} \circ \mathrm{pad}_{\text{TP}\times\nu} \circ
    \Pi_{\text{CP}} \circ \mathrm{pad}_1$, with $\Pi_{\text{CP}}$ the interleaved
    context-parallel slice and $\Pi_{\text{SP}}$ the local sequence-parallel shard
    (Appendix~\ref{app:align}). \\
$\nu$ & integer & Padding granularity of the packed sequence. \\
$\gS$ & mode
  & Router mode: free selection, replayed selection without gradient, or replayed
    selection with gradient. \\
\bottomrule
\end{tabularx}
\end{table}

\begin{table}[H]
\tablesize
\centering
\caption{RL objective and diagnostics (Sections~\ref{sec:other-stability},
\ref{sec:critic}).}
\label{tab:notation-rl}
\begin{tabularx}{\linewidth}{@{}l l X@{}}
\toprule
\textbf{Symbol} & \textbf{Type} & \textbf{Meaning} \\
\midrule
$s_j,\ a_j,\ x_j$ & state, action, token
  & Conditioning prefix, emitted token, and stream token at position $j$. \\
$r_j(\theta)$ & scalar
  & PPO importance ratio (Equation~\ref{eq:ppo-ratio}). Distinct from the per-position
    reward $\hat r_j$. \\
$\hat r_j$ & scalar
  & Reward credited at position $j$; nonzero only on a trajectory's final token
    (Section~\ref{sec:reward-def}). \\
$\varepsilon$ & scalar
  & PPO clip ratio, $0.2$. Distinct from the router-perturbation bound $\epsilon$. \\
$\hat A_j,\ \hat R_j,\ \delta_j$ & scalars
  & GAE advantage, return, and TD residual (Equation~\ref{eq:ppo_updates}). \\
$\gamma,\ \lambda$ & scalars
  & Discount and GAE decay; both $1.0$ here. $\gamma$ is distinct from the margin
    $\gamma^{\ell}_j$. \\
$\gL^{\text{PPO}},\ \gL^{V}$ & scalars & Policy and value objectives. \\
$\mathrm{EV}$ & scalar & Critic explained variance (Equation~\ref{eq:ev}). \\
$\Delta_j$ & scalar
  & Per-token train--inference log-probability discrepancy
    (Equation~\ref{eq:delta-decomp}). \\
$|\Delta \log p|$ & scalar
  & Mask-weighted mean $|\Delta_j|$ over the loss region (Equation~\ref{eq:tis}). \\
$S$ & integer & Number of unit tests in a task's verifier (Section~\ref{sec:reward}). \\
\bottomrule
\end{tabularx}
\end{table}

\section{Asynchrony, version skew, and routing alignment}
\label{app:async}

This appendix makes precise what one-step asynchrony means for the quantities of
Section~\ref{sec:stability}, and why routing alignment must be defined against a specific
version rather than against the policy in the abstract.

\subsection{The one-step-ahead loop}

Training and inference occupy disjoint devices, and the rollout for step $t{+}1$ generates
while step $t$ trains. Writing $\gB_t$ for the batch consumed by update $t$, the loop
maintains
\begin{equation}
  \label{eq:async-loop}
  \gB_t \sim \pi^{\mathrm{r}}_{t-\sigma},
  \qquad
  \theta_t = \theta_{t-1} - \eta_\theta \nabla_\theta \gL^{\text{PPO}}(\theta_{t-1}; \gB_t),
  \qquad \sigma = 1 ,
\end{equation}
with the publication of $\theta_{t}$ synchronized against in-flight generation so that no
request ever spans two versions. The staleness $\sigma$ is therefore a constant of the
production configuration rather than a random variable, since every sample in $\gB_t$ was
drawn under exactly $\pi_{t-1}$. A fully asynchronous path with unbounded $\sigma$ exists
but was not used for the 122B campaigns.

\subsection{Three distinct gaps, one measured quantity}

Because version and implementation are independent axes, the discrepancy $\Delta_j$ that
Section~\ref{sec:mismatch-effect} measures decomposes into terms with different causes and
different remedies. For a loss-bearing position $j$,
\begin{align}
  \label{eq:delta-decomp}
  \Delta_j
  &= \underbrace{\log \pi^{\mathrm{t}}_{t}(x_j \mid \cdot)
               - \log \pi^{\mathrm{t}}_{t-1}(x_j \mid \cdot)}_{
       \text{(a) version skew, genuine learning}}
   + \underbrace{\log \pi^{\mathrm{t}}_{t-1}(x_j \mid \cdot)
               - \log \pi^{\mathrm{r}}_{t-1}(x_j \mid \cdot)}_{
       \text{(b) implementation gap, to be eliminated}} , \\
  \label{eq:engine-gap}
  \text{(b)}
  &= \underbrace{\big[\text{token drift}\big]}_{\text{\TITO}, \;
       \text{Equation~\ref{eq:tito-identity}}}
   + \underbrace{\big[\text{routing divergence}\big]}_{\Rthree, \;
       \text{Equation~\ref{eq:r3-identity}}}
   + \underbrace{\big[\text{kernel numerics}\big]}_{\text{irreducible here}} .
\end{align}
Three consequences follow, and they explain the shape of every mismatch curve in this
report.

\paragraph{Only the second term is a defect.}
Term (a) is the distance the policy legitimately travelled during one update. It is what
PPO's clip exists to bound, and it should be nonzero. Term (b) is bookkeeping error,
namely the same weights $\theta_{t-1}$ disagreeing with themselves across two
implementations, and \TITO together with \Rthree target it exclusively. This is why the
slow upward drift of $|\Delta \log p|$ over a long run is not a regression, as
Section~\ref{sec:stability-indicators} notes: as the policy improves, term (a) grows while
term (b) stays flat.

\paragraph{The behaviour policy is evaluated on the training side by choice.}
The PPO denominator is recomputed by the trainer rather than read back from the sampler,
so Equation~\ref{eq:ppo-ratio} instantiates as
\begin{equation}
  \label{eq:ppo-ratio-async}
  r_j(\theta) = \frac{\pi^{\mathrm{t}}_{t}(x_j \mid x_{<j}, \sI_j)}
                     {\pi^{\mathrm{t}}_{t-1}(x_j \mid x_{<j}, \sI_j)},
  \qquad
  \sI_j = \tR[j, :, :] \;\text{ for both numerator and denominator.}
\end{equation}
Both factors are evaluated by the same implementation on the same routing, so term (b)
cancels from the ratio provided $\sI_j$ is held fixed across the two passes, which is
exactly what \Rthree guarantees and what Algorithm~\ref{alg:r3} schedules, with the
gradient-free replay supplying the denominator and the gradient-carrying replay the
numerator. Sampler log-probabilities $\vq$ are then carried for diagnostics alone.

\paragraph{Routing must be pinned to the generating version.}
The recorded tensor $\tR$ is a property of $\pi^{\mathrm{r}}_{t-1}$, the version that
produced the tokens, and replaying it while computing gradients for $\theta_t$ is
deliberate:
\begin{equation}
  \label{eq:routing-version}
  \sI^{\ell}_j \gets \tR[j,\ell,:] = \sI^{\mathrm{r},\ell}_j\big(\pi_{t-1}\big),
  \qquad
  g^{\ell}_{j,e} \gets \big[\vs^{\ell}_j(\theta_t)\big]_e ,
\end{equation}
so that selection is frozen at the generating version while gating is evaluated at the
current one. The asymmetry is what makes the router trainable under replay: gradients
reach $\vs^{\ell}_j(\theta_t)$ through $g^{\ell}_{j,e}$, whereas the discrete
$\operatorname{TopK}_k$, whose sensitivity to $\gamma^{\ell}_t \le 2\epsilon$ caused the
problem in the first place, is removed from the graph. Under $\sigma = 1$ the frozen
selection is one update stale, which is the same staleness PPO's ratio already corrects
for. Under large $\sigma$ it would not be, and this is the mechanism by which the
staleness studies of Section~\ref{sec:limitations} report baselines collapsing at
$\sigma = 8$ while combinations based on \Rthree survive.

\subsection{Routing alignment across a version boundary}
\label{app:align}

Equation~\ref{eq:routing-version} is a statement about one position. Making it hold for
every position of a stitched multi-turn trajectory is the actual engineering content,
because $\tR$ is captured per turn by the replica holding $\pi_{t-1}$, while the loss is
computed over a single concatenated stream sharded across ranks by the trainer holding
$\theta_t$. Algorithm~\ref{alg:align} states the invariant chain connecting them.

\begin{algobox}{Routing alignment from capture under $\pi_{t-1}$ to replay under
$\theta_t$. Every step is index-preserving, and a violation anywhere silently trains the
wrong sub-network, so each is asserted rather than assumed.}{alg:align}
\Require per-turn captures $\{(\vp_i, \va_i, \tR_i)\}_{i=1}^{T}$ from $\pi^{\mathrm{r}}_{t-1}$
\Statex \textbf{Per-turn offset.} $\tR_i$ holds $\rho_i = |\vp_i| + |\va_i| - 1$ records,
        where record $j$ predicts position $j{+}1$, so a turn's last token contributes
        none.
\Statex \textbf{Stitch-consistent concatenation.} Records are appended under the same case
        decision, Equations~\ref{eq:case-strict} to \ref{eq:case-split}, that built
        $(\vx, \vm, \vq)$ and never independently, so $\tR[j,\ell,:] =
        \sI^{\mathrm{r},\ell}_j$ for the same $j$ indexing $(x_j, m_j, q_j)$.
\Statex \textbf{Gap repair confined to masked positions.}
        $\tR[j] \gets \tR[j{-}1]$ is permitted iff $m_{j+1} = 0$ and otherwise aborts, per
        Equation~\ref{eq:placeholder}, so no replayed record influencing a gradient is
        synthetic.
\Statex \textbf{Length contract.} \textbf{assert} $\rho = N - 1$ per sample; a mismatch
        means the preceding invariants disagree and is fatal.
\Statex \textbf{Identical sharding.} Apply
        $\Pi = \Pi_{\text{SP}} \circ \mathrm{pad}_{\text{TP}\times\nu} \circ
        \Pi_{\text{CP}} \circ \mathrm{pad}_1$ to $\tR$, in the same composition and the
        same order as for $\vx$. Any deviation permutes records relative to tokens.
\Statex \textbf{Version-split substitution.} In the forward pass under $\theta_t$,
        selection comes from $\tR$ at version $t{-}1$ without gradient and gating from
        $\vs^{\ell}_j(\theta_t)$ at version $t$ with gradient, per
        Equation~\ref{eq:routing-version}.
\Statex \textbf{Cursor discipline.} The router advances a forward cursor for the
        $\pi^{\mathrm{t}}_{t-1}$ pass and a backward cursor for the $\theta_t$ update,
        because activation recomputation re-runs each layer's forward pass during the
        backward pass, so one cursor would be consumed twice per layer.
\Statex \textbf{Scope.} Reference and critic passes select freely, since neither needs
        behavioural fidelity to $\pi^{\mathrm{r}}_{t-1}$, and the critic's distinct
        batching would violate identical sharding on shared buffers.
\end{algobox}

The first five invariants concern indexing and would be required even in a synchronous
loop, whereas the last three are where the version boundary appears explicitly. What is
not claimed is that alignment makes $\pi^{\mathrm{t}}_{t-1}$ bit-identical to
$\pi^{\mathrm{r}}_{t-1}$, since the kernel-numerics term of Equation~\ref{eq:engine-gap}
survives. It makes the two agree on which sub-network is differentiated, and that is the
term scaling with $L$, hence the one that matters at 48 layers of top-8-of-256 routing.

\subsection{Version bookkeeping in the critic path}
The critic introduces a second version axis, and the two must not be conflated. Each step
the critic updates first and then supplies its pre-update values to the actor:
\begin{equation}
  \label{eq:critic-version}
  \phi_t = \phi_{t-1} - \eta_\phi \nabla_\phi \gL^{V}(\phi_{t-1}; \gB_t),
  \qquad
  V_{\mathrm{old}} \equiv V_{\phi_{t-1}}
  \quad\text{in both } \hat A_j \text{ and the value clip.}
\end{equation}
Anchoring GAE and the $\pm 0.2$ value clip to $V_{\phi_{t-1}}$ rather than to the
in-flight $V_{\phi_t}$ keeps the clip a trust region around a fixed function. The critic
never replays routing, so its forward pass runs at $(\phi_{t-1}, \operatorname{TopK}_k)$,
and its target is a supervised regression on returns for which behavioural fidelity to
$\pi^{\mathrm{r}}_{t-1}$ is not required.

Finally, the frozen reference $\pi_{\mathrm{ref}}$ carries no version subscript because it
does not advance. It also routes with its own $\operatorname{TopK}_k$ rather than with
$\tR$, which is why
\begin{equation}
  \label{eq:kl-async}
  \KL\!\big(\pi^{\mathrm{t}}_{0} \,\big\|\, \pi_{\mathrm{ref}}\big) > 0
  \quad\text{at initialization, even though } \theta_0 = \theta_{\mathrm{ref}} ,
\end{equation}
and why any assertion of a vanishing initial divergence must be disabled under \Rthree, as
Section~\ref{sec:r3} notes. The two policies differ not in weights but in routing source.
\section{Case Study}
\label{app:case-study}

\begin{figure}[t]
  \centering
  \includegraphics[width=\linewidth]{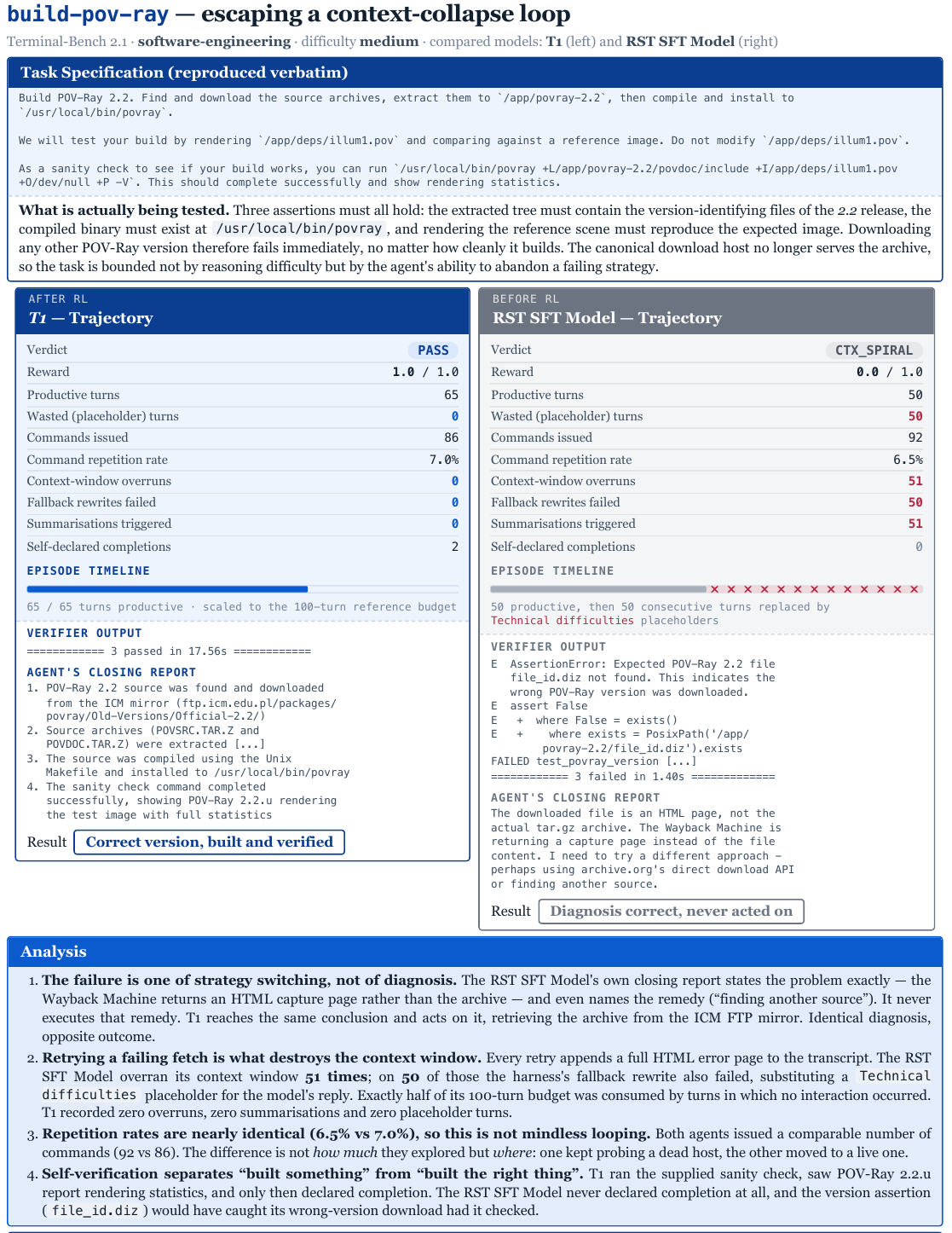}
  \caption{\textbf{Escaping a context-collapse loop.}
  The task specification and the three assertions it enforces are reproduced at the top;
  the panels compare T1 \figleft{} with the RST-SFT model \figright{} on trajectory
  statistics, episode timeline, verifier output, and the agent's own closing report.
  T1 resolves the task in 65 productive turns with no context overruns, while the SFT
  model spends 50 of its 100 turns on placeholder replies after 51 overruns and fails the
  version assertion (\code{file\_id.diz}) by downloading the wrong POV-Ray release.}
  \label{fig:case_ray}
\end{figure}

\begin{figure}[t]
  \centering
  \includegraphics[width=\linewidth]{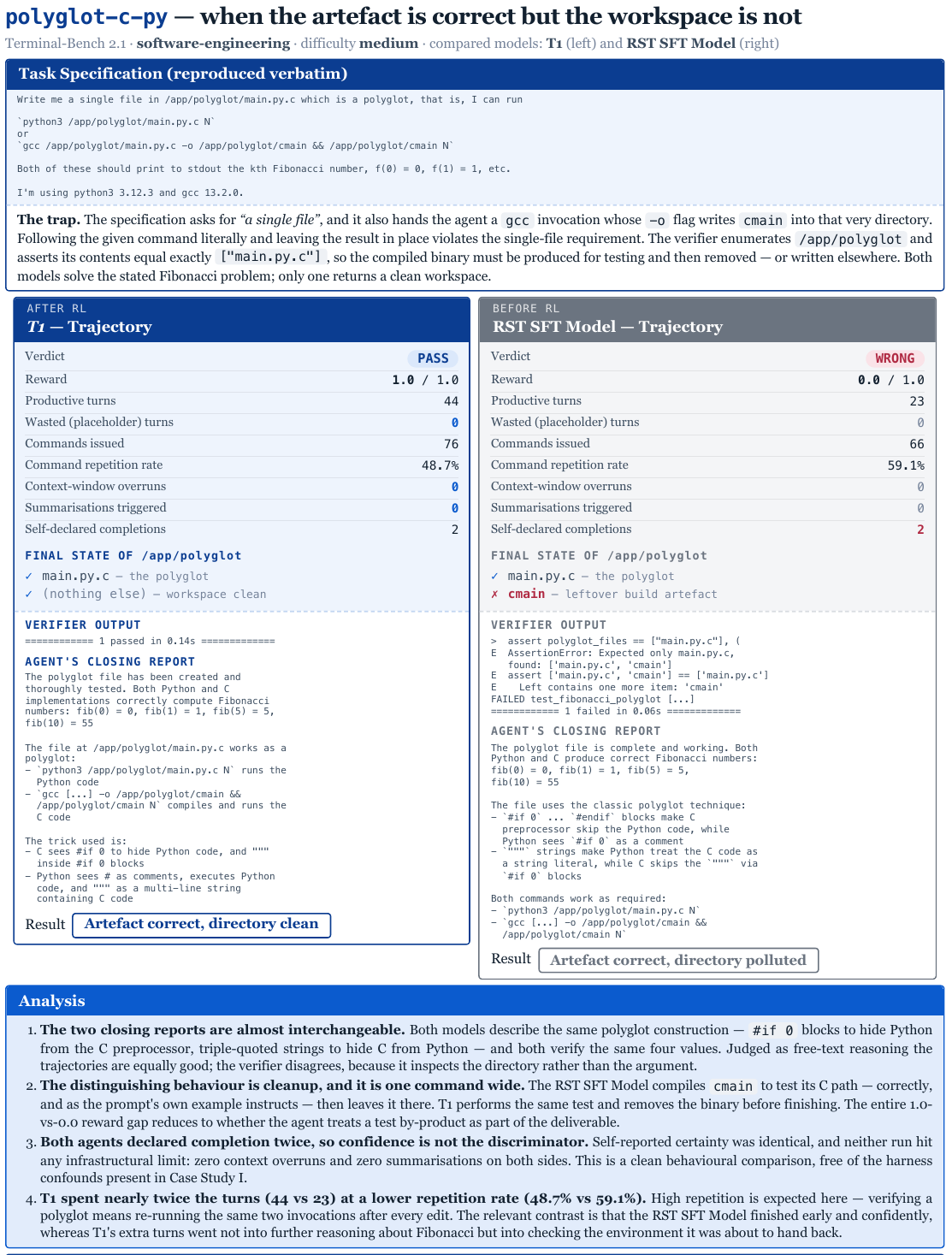}
  \caption{\textbf{Correct artefact in a polluted
  workspace.} Layout follows Figure~\ref{fig:case_ray}, with the final contents of
  \code{/app/polyglot} shown for each model. Both agents produce a working
  Python/C polyglot and report the same verified Fibonacci values, but the verifier
  requires the directory to hold exactly \code{main.py.c}: the RST-SFT model leaves the
  compiled \code{cmain} binary in place and scores 0.0, whereas T1 removes it after
  testing and scores 1.0. Neither run triggers a context overrun or a summarization.}
  \label{fig:case_pov}
\end{figure}

Figures~\ref{fig:case_ray} and~\ref{fig:case_pov} place T1 next to its own RST-SFT
initialization on two \tbtwoone software-engineering tasks of medium difficulty that
T1 resolves and the SFT model does not. Both pairs run the same task specification and
the same verifier, so the reward difference is attributable to behaviour rather than to
the problem statement.

\begin{itemize}
\item \textbf{\code{build-pov-ray} (Figure~\ref{fig:case_ray}): abandoning a failing
strategy.} The canonical download host no longer serves the POV-Ray 2.2 archive. The SFT
model diagnoses this correctly in its own closing report, naming the remedy, yet keeps
retrying the dead host; each retry appends an HTML error page, producing 51 context
overruns and 50 turns replaced by harness placeholders, so half its budget carries no
interaction. T1 reaches the same diagnosis, switches to a live FTP mirror, and passes
all three assertions in 65 turns with zero overruns. Command counts (86 vs.\ 92) and
repetition rates (7.0\% vs.\ 6.5\%) are comparable: what differs is where the search went,
not how much of it there was.

\item \textbf{\code{polyglot-c-py} (Figure~\ref{fig:case_pov}): satisfying the unstated
requirement.} Both models write a correct Python/C polyglot and verify the same Fibonacci
values, and their closing reports are nearly interchangeable. The verifier, however,
asserts that \code{/app/polyglot} contains exactly \code{main.py.c}. The SFT model
compiles the test binary as the prompt's own example instructs and leaves it behind;
T1 runs the same test and deletes it. Neither run approaches any harness limit, so the
entire 1.0-vs-0.0 gap reduces to one cleanup command. T1 spends its extra turns
(44 vs.\ 23) inspecting the environment it is about to return rather than reasoning
further about the algorithm.
\end{itemize}

Both cases show RL changing what the policy does with a conclusion it can already reach:
discarding an exhausted hypothesis, and treating the final machine state as part of the
deliverable. The two runs differ in harness limits (SFT 56k/8{,}192 tokens and 100 turns;
T1 86k/32{,}768 tokens and 500 turns), which we note as a caveat, although T1
finished inside the SFT budget in both cases and neither recorded failure cause is one
that additional context would have repaired.

\end{document}